%% file: acl_latex_copy.tex
\documentclass[11pt]{article}

\usepackage[preprint]{acl}

\usepackage{times}
\usepackage{latexsym}

\usepackage[T1]{fontenc}

\usepackage[utf8]{inputenc}

\usepackage{microtype}

\usepackage{inconsolata}

\usepackage{graphicx}

\usepackage[table]{xcolor}
\definecolor{lightgreen}{rgb}{0.88,1,0.88}
\definecolor{lightred}{rgb}{1,0.88,0.88}

\usepackage{enumitem}
\usepackage{pgfplots}
\usepackage{tabularx}
\usepackage{caption}
\usepackage{subcaption}
\usepackage{bbm}
\usepackage{url}
\usepackage{amsmath} 
\usepackage{pifont}
\usepackage{arydshln}
\usepackage{makecell}
\usepackage{booktabs}
\usepackage{multirow} 
\usepackage{algorithm}
\usepackage{algorithmic}
\usepackage{amssymb}
\usepackage{xcolor}
\usepackage{tcolorbox}
\usepackage{bbding}
\usepackage{siunitx}
\usepackage{newtxmath}

\definecolor{BlueRow}{HTML}{E6F4FF}
\newcommand{\highlightrow}{\rowcolor{BlueRow}}

\title{Let the Model Distribute Its Doubt: Confidence Estimation through\\Verbalized Probability Distribution}

\author{
 \textbf{Ante Wang}\textnormal{\textsuperscript{$\diamondsuit$}},
 \textbf{Weizhi Ma}\textnormal{\textsuperscript{$\diamondsuit$}} \textnormal{and}
 \textbf{Yang Liu}\textnormal{\textsuperscript{$\clubsuit$,$\diamondsuit$}}
\\
 \textsuperscript{$\clubsuit$}Dept. of Comp. Sci. \& Tech., Institute for AI, Tsinghua University, Beijing, China\\
 \textsuperscript{$\diamondsuit$}Institute for AI Industry Research (AIR), Tsinghua University, Beijing, China
}

\begin{document}
\maketitle
\begin{abstract}
Knowing the reliability of a model's response is essential in practical applications. Given the strong generation capabilities of large language models (LLMs), research has focused on generating verbalized confidence. This approach is further enhanced by integrating chain-of-thought reasoning, which provides logical and transparent estimates. However, how reasoning strategies affect the estimated confidence remains under-explored. In this work, we demonstrate that predicting a verbalized probability distribution effectively promotes reasoning for confidence estimation. It requires an LLM to consider all possible answers rather than relying on a single guess, and the requirement of producing a distribution elicits more careful confidence assignment. We conduct systematic experiments comparing different verbalization-based methods across multiple LLMs and tasks. Our method consistently shows advantages, whether in the simple prompting setup or after optimization via reinforcement learning (RL). Notably, it achieves higher reasoning efficacy during inference-time scaling, saving nearly 6$\times$ the computation to reach the best Brier score of the strongest baseline on MMLU-Pro. Additionally, we reveal its limitations on specific tasks and discuss possible solutions for broader applicability.
\end{abstract}

\section{Introduction}

Despite significant progress on various tasks, Large Language Models (LLMs,~\citealt{achiam2023gpt,jiang2023mistral7b,yang2025qwen3,liu2024deepseek,guo2025deepseek}) still inevitably make errors, preventing their deployment in high-stakes applications such as healthcare, law, and finance~\cite{jiang2012calibrating,bojarski2016end}. A promising solution to this limitation is the development of well-calibrated confidence estimates, which would allow LLMs to differentiate between correct and incorrect answers and produce confidence scores that match their true empirical accuracy. This capability is essential for building trustworthy AI systems. For instance, if an LLM expresses low confidence in a clinical diagnosis, it could automatically trigger a more careful analysis by a human physician.

Traditional methods typically focus on classification tasks and rely on scores produced by a classification head~\cite{guo2017calibration,kumar2019verified}. 
This approach can be naturally applied to LLMs by aggregating their generated token probabilities~\cite{kadavath2022language}.
However, research~\cite{tian2023just} demonstrates that modern LLMs, particularly those fine-tuned with Reinforcement Learning from Human Feedback (RLHF,~\citealt{ouyang2022training}), may sacrifice well-calibrated predictions to better adhere to user instructions. To address this, verbalized confidence~\cite{tian2023just,yangverbalized}, where a model expresses its uncertainty directly in natural language, has been proposed. Nevertheless, empirical analyses~\cite{xiongcan,li2024think} reveal that this method can suffer from overconfidence and its reliability is often dependent on model scale.

Inspired by the success of chain-of-thought reasoning~\cite{wei2022chain}, recent work~\cite{zhang2025cot,yoon2025reasoning,damani2025beyond} has shown that LLMs can reason about their own uncertainty to produce more accurate confidence estimates. These methods prompt the model to analyze its own prediction before providing a confidence score. However, it remains unclear what constitutes effective reasoning content for this purpose. To address this gap, we investigate predicting verbalized probability distributions.
This approach forces LLMs to express their entire belief state, contextualizing the likelihood of their primary prediction by considering all alternatives, thereby mitigating the short-sightedness that leads to overconfidence.
Besides, in this way, the model is implicitly asked to match its confidence to the empirical frequency of correctness, aligning with the definition of calibration.

To implement this, we directly prompt the LLMs to predict all possible answers and assign confidence values to form a probability distribution.
Nevertheless, due to model capacity and the nature of some questions, it can be difficult to enumerate all possible answers correctly. To address this, we include a ``None of the above'' option to aggregate the probability of low-likelihood candidates.
Furthermore, building on prior work~\cite{damani2025beyond}, we explore optimizing the reasoning process via reinforcement learning (RL) by incorporating an additional term in the reward function to encourage the generation of proper probability distributions.

\begin{figure*}
\centering
\includegraphics[width=0.98\textwidth]{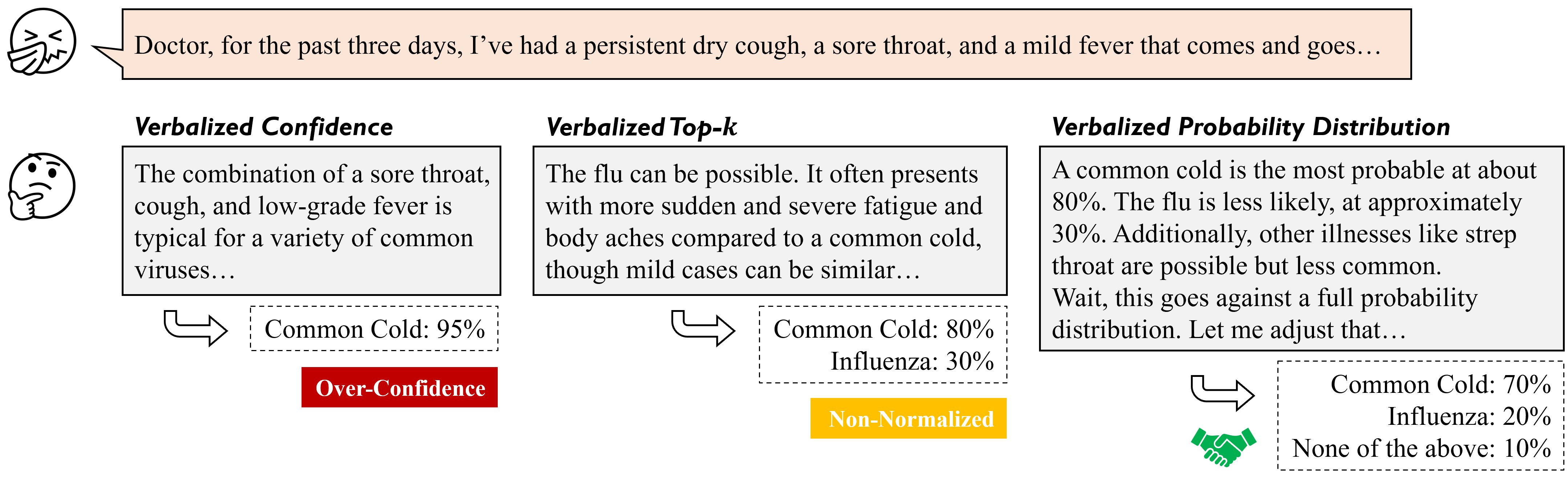}
\caption{An example for illustrating the difference in the three verbalization-based methods.}
\label{fig:method}
\end{figure*}

We evaluate our approach on datasets with either closed-set or open-set answer spaces across various LLMs. For closed-set tasks, we use MedQA~\cite{jin2021disease}, MMLU-Pro~\cite{wang2024mmlu}, and MedXpertQA~\cite{zuomedxpertqa}, which are all multiple-choice tasks with limited options. For open-set tasks, we adopt HotpotQA~\cite{yang2018hotpotqa} and MedCaseReasoning~\cite{wu2025medcasereasoning}, where the set of possible answers is too broad to enumerate.
Results indicate that, using only training-free prompting, our method achieves competitive performance on simpler datasets and significantly outperforms baselines on more challenging ones. Experiments with RL show that our method converges faster and demonstrates robust performance on out-of-domain datasets. We further analyze how probability distribution prediction affects the reasoning process and validate the efficiency of our method. Finally, we discuss the limitations of our approach for tasks where the answer space is difficult to predict and suggest possible ways to address them.

\section{Preliminary}

Formally, given a question $q$, an LLM $\pi_\theta$ predicts an answer $a$ along with a corresponding confidence $c$. In this study, we focus primarily on the prediction of $c$ using verbalization-based approaches (\S \ref{sec:base}). This process can be naturally enhanced via chain-of-thought (CoT) reasoning and further optimized through RL (\S \ref{sec:rlcr}).

\subsection{Verbalization-Based Methods}
\label{sec:base} 
This paradigm uses natural language reasoning to articulate confidence values. It effectively harnesses the reasoning capabilities of the latest LLMs and is applicable to black-box models where token probabilities are inaccessible.
In this study, we introduce two prevalent methods in this field, both proposed in \cite{tian2023just}.

\begin{itemize}[leftmargin=*]
\item \textbf{Verbalized Confidence} directly prompts an LLM to reason about both an answer and its confidence score. The confidence can be either numerical probabilities or another linguistic expression of uncertainty. Both implementations have been proven effective in prior research; thus, we adopt numerical confidence in this study for its convenience.
\item \textbf{Verbalized Top-$k$}, in contrast, prompts the model to produce $k$ guesses alongside a probability for each. The highest-probability prediction is then selected as the final output. It is inspired by research in human psychology~\cite{lord1984considering,mussweiler2000overcoming} and it has been validated that this can effectively mitigate the overconfidence issue.
\end{itemize}

Though this approach does not necessarily employ CoT reasoning, it has been shown to help achieve more precise confidence estimation on challenging tasks~\cite{yoon2025reasoning,devic2025trace,mei2025reasoning}.

\subsection{RL for Confidence Estimation}
\label{sec:rlcr}
Recent advances have demonstrated the efficacy of RL in refining reasoning processes.  
By optimizing for a goal quantified as a reward, RL allows LLMs to automatically explore reasoning strategies on-policy, a technique that has proven to yield significant improvements on reasoning-intensive tasks~\cite{yang2025qwen3,guo2025deepseek}.  

For confidence estimation, \citet{damani2025beyond} proposed reinforcement learning with calibration rewards (RLCR), which optimizes the reasoning process \(\tau\) to predict precise verbalized confidence. The confidence is generated as \(\tau, a, c \sim \pi_\theta(q, p)\), where \(p\) is the input prompt.  
Given the ground truth \(a^*\) for a question, RLCR adopts a reward function of \(r = y - (c - y)^2\), where \(y = \mathbbm{1}(a, a^*)\) is the answer correctness and \((c - y)^2\) denotes the Brier score. This ensures that confidence is calibrated without sacrificing answer accuracy.  

\section{Verbalized Probability Distribution}
In this work, we propose Verbalized Probability Distribution, a method that prompts LLMs to reason about a probability distribution over the entire space of possible answers, thereby encouraging comprehensive confidence quantification.  
Formally, we prompt the LLM to predict all possible answers $\mathbf{a}=a_1,\dots,a_n$ and assign confidence scores $\mathbf{c}=c_1,\dots,c_n$ to each:
\[\tau, \mathbf{a}, \mathbf{c} \sim \pi_\theta(q, p^*),\]
where $p^*$ is the prompt, with the specific instruction that the sum of $\mathbf{c}$ should equal $1$.  
We take the prediction with the highest confidence as the final answer $a_i$, where $i = \arg\max(\mathbf{c})$.  

For open-ended questions where enumerating all answers is infeasible, applying a probability mass of $1$ only to the available answers is not reasonable. To address this, we include a ``None of the above'' option to aggregate the probability of all low-likelihood candidates.  

An alternative implementation is to manually normalize the confidence scores. However, we believe that LLMs can better control confidence assignment for each answer through this additional reasoning process, avoiding the unexpected results through simple mechanical normalization.  

Furthermore, we investigate whether RL can further enhance confidence estimation by optimizing a model's verbalized probability distribution.  
We follow RLCR~\cite{damani2025beyond} in designing the reward function but adopt a strict penalty for outputs that do not constitute a valid probability distribution. That is, if $\sum \mathbf{c} \neq 1$, we directly assign a score of $r = -1$.

\paragraph{Discussion}
A comparison of the different verbalization-based methods is illustrated in Figure~\ref{fig:method}.  
Verbalized Confidence can be blind to a single prediction without considering alternatives, leading to an overconfidence issue.  
Verbalized Top-$k$ considers a fixed number of answers, mitigating the overconfidence issue; however, it may overlook how confidence is distributed among alternatives, resulting in infeasible confidence quantification (e.g., the sum of the confidence scores for the two answers in Figure~\ref{fig:method} reaches $110\% > 100\%$).  
In contrast, while our method shares conceptual similarities with Verbalized Top-$k$ in terms of considering alternatives, it is distinguished by its emphasis on reasoning about the entire belief state and producing a reasonable confidence assignment that satisfies the requirements of a probability distribution.  
In \S \ref{sec:exp}, we investigate whether prompting LLMs to reason about these improvements can elicit better confidence estimation.

\section{Experiments}
\label{sec:exp}

\input{tables/unify_train_free_verb}

\input{figures/calibration}

\subsection{Setup}

\paragraph{Datasets}
We focus on tasks that require reasoning to predict the best answer. To encompass various scenarios, the tasks in this work are categorized based on whether the answer space is closed-set or open-set.
For \textbf{closed-set} tasks, we evaluate three multiple-choice tasks requiring reasoning: MedQA~\cite{jin2021disease}, MMLU-Pro~\cite{wang2024mmlu}, and MedXpertQA~\cite{zuomedxpertqa}. For \textbf{open-set} tasks, we evaluate two open-ended generation tasks: HotpotQA~\cite{yang2018hotpotqa} and MedCaseReasoning~\cite{wu2025medcasereasoning}.
The medical domain is selected as a suitable and important case study due to its practical relevance and reasoning-intensive nature.
Detailed descriptions of these datasets can be found in \S\ref{app:dataset}.

\paragraph{Evaluation metrics}
Following established literature, we evaluate confidence estimation performance using the Area Under the Receiver Operating Characteristic curve (\textbf{AUROC}), the Expected Calibration Error (\textbf{ECE}), and the \textbf{Brier} Score. These metrics provide complementary views of model performance: AUROC measures discrimination between positive and negative classes; ECE quantifies calibration relative to empirical accuracy; and the Brier Score evaluates both aspects simultaneously.
Besides, good confidence estimation should not come at the cost of answer accuracy. Therefore, we also report \textbf{Accuracy} to verify this.
Further details on these metrics are provided in \S\ref{app:metric}.

\paragraph{Implementation details}
We summarize the prompts used for different verbalization-based approaches in \S\ref{app:prompts}. Following \citet{tian2023just} and \citet{tao2024trust}, we set $k=2$ in Verbalized Top-$k$ across all experiments. Our empirical results (Table \ref{tab:top_k}) demonstrate that increasing $k$ does not lead to substantial improvement.
We experiment with various LLMs, as detailed in \S\ref{app:model}.
We adopt GRPO~\cite{shao2024deepseekmath} as our RL algorithm. Following \cite{damani2025beyond,bereketuncalibrated}, we remove the division by the standard deviation in the advantage calculation, as this has been shown to mitigate poor calibration. The detailed training configuration is provided in \S\ref{app:train}.
To accurately compute the answer correctness during both testing and RL training, we use exact match for closed-set tasks and LLM-as-a-Judge~\cite{zheng2023judging} for open-set tasks to ensure more precise assessment. Evaluation prompts are provided in \S\ref{app:prompts}.

\input{tables/unify_train_based_verb}

\input{figures/train_step}

\subsection{Main Results}
\label{sec:main}
Given the black-box nature of current LLMs and the high cost of training large-scale models, training-free prompting offers greater general applicability. In contrast, RL training allows us to mitigate inherent biases within LLMs, thereby better revealing the potential of each method.
In this study, we compare different verbalization-based approaches in both training-free and RL-training setups, as shown in Tables~\ref{tab:unify_train_free_verb} and \ref{tab:unify_train_based_verb}, respectively. Our conclusions are as follow.

\paragraph{Better results through direct prompting}
Among the verbalization-based approaches in Table~\ref{tab:unify_train_free_verb}, Verbalized Confidence displays a high degree of over-confidence, resulting in worse calibration (higher Brier and ECE scores). However, this issue is mitigated when using more capable LLMs, a finding that aligns with prior research.
Verbalized Top-$k$ accelerates the mitigation of over-confidence and thus consistently improves calibration performance. Nonetheless, its effectiveness remains limited for smaller LLMs and more challenging tasks.
In contrast, Verbalized Probability Distribution demonstrates significantly better calibration on both closed-set and open-set tasks across all LLMs. Besides, it maintains answer accuracy scores and achieves noticeably higher AUROC scores, indicating better discrimination ability. Crucially, smaller LLMs gain more benefits from reasoning verbalized probability distributions, highlighting the practical utility of this method. The calibration curve in Figure~\ref{fig:calib_curve} further illustrates the strong alignment between predicted confidence and empirical accuracy.

\input{tables/ablation_study}

\input{figures/sampling}

\paragraph{The advantage is preserved after RL training}
Through extensive experience gained during RL, an LLM may learn an optimal strategy that works for various prompts. We observe that the performance improves significantly for all methods, and the gap between different verbalization-based methods narrows. However, the overall trends remain consistent with the training-free evaluation. The Verbalized Probability Distribution method maintains its advantage across different tasks, especially the more challenging ones. This is likely due to its better initialization, which results in faster and smoother convergence, as shown in Figure~\ref{fig:train_step}. 
This demonstrates that the choice of prompt used for training confidence estimation is crucial. 
It appears difficult for an LLM to independently discover the optimal reasoning strategy through RL alone.
This finding raises a question: can a longer prompt with more fine-grained guidance lead to even better performance?
We implemented the long prompt provided in \cite{damani2025beyond}. The results, shown in Table~\ref{tab:rlcr_prompt}, validate that the performance gains stem from the requirement to predict a distribution, rather than from more complex instructions.
Another interesting question for future work is whether larger LLMs with stronger exploration capabilities could mitigate this issue, though testing this would require computational resources beyond our current means.

\input{figures/answer_space}

\subsection{How Verbalized Probability Distribution Enhances Reasoning}
\label{sec:analysis}

This section provides a deeper analysis of how the Verbalized Probability Distribution enables better reasoning for confidence estimation, examining it from the perspectives of prompt ablation, inference efficiency, and answer space prediction.

\paragraph{What constitues more effective reasoning}
Table~\ref{tab:ablation_study} presents the ablation study of our method. First, removing the ``None of the above'' option from the input prompt causes a significant performance drop across all metrics, especially for smaller LLMs. This occurs because the models cannot enumerate all possible alternatives before normalizing the distribution, leading to unreasonable confidence assignments. Next, removing the normalization requirement degrades the method to an enhanced version of Verbalized Top-$k$ (where $k$ covers all possible answers, including ``None of the above'' for open-set tasks). This has a different effect on the AUROC score across tasks; nonetheless, it notably harms calibration, underscoring the importance of normalization for those metrics. An alternative implementation, manually normalizing the probabilities instead of having the LLM do so, sometimes performs even better in terms of calibration, further confirming the value of normalization. However, as shown in Tables \ref{tab:ablation_multi_choice} and \ref{tab:ablation_open_end}, this manual normalization is effective only on challenging tasks (e.g., MedXpertQA and MedCaseReasoning) where LLMs struggle with confidence estimation. When applied in this mechanical manner without autogenic reasoning from the LLMs, it obviously hurts AUROC scores especially on open-set tasks.

\paragraph{Efficiency analyses through inference time scaling}
Different verbalization-based methods incur different token consumption. To fairly compare these methods regarding reasoning efficiency, we investigate combining these approaches with inference-time scaling, i.e., leveraging answer consistency across multiple sampled trajectories~\cite{wang2024self,xiongcan}.
We select MMLU-Pro as the test set because it covers wider domains and its closed-set nature enables accurate answer clustering. As shown in Figure~\ref{fig:consistency_mmlu}, we sample at most 16 reasoning trajectories using a temperature of 0.8. We follow \citet{wang2024self} in using answer frequency as the confidence score for our baseline. For the verbalization-based approach, the predicted confidence scores are used as weights in a weighted aggregation~\cite{xiongcan}. Specifically, for each unique answer, we sum its corresponding confidence scores across all predictions. The answer with the highest aggregate score is selected, and its final confidence score is obtained by normalizing this sum.
Results show that when the computational budget is increased, all these methods generally produce better confidence scores. Verbalized Confidence performs even worse than the Baseline. Verbalized Top-$k$ exhibits an abnormal trend in ECE, possibly because of its non-normalized predictions. In contrast, Verbalized Probability Distribution demonstrates better sample efficiency, consistently achieving better scores across different metrics with relatively lower token costs.

\paragraph{Analysis of the prediction of the answer space}
Predicting the answer space is important for our method in open-set tasks. To analyze this, we divided the predictions from HotpotQA into 10 bins based on confidence scores and examined several aspects of how the answer space is structured.
As shown in Figure~\ref{fig:answer_space}, when LLMs exhibit lower confidence, they tend to predict more alternative answers and allocate higher probability to ``None of the above.'' This behavior is expected, as it distributes uncertainty across competing or indecisive alternatives, leading to better-calibrated confidence. However, this trend is not entirely stable; we observe sharp fluctuations in these metrics, most notably a sharp drop in the probability mass assigned to ``None of the above'' within the 0.3-0.4 confidence interval for both LLMs. Nevertheless, after RL training, stability improves substantially, with all metrics showing much smoother trends as confidence increases.
Due to its stronger capabilities, GPT-4.1 achieves higher recall of the gold answer while predicting fewer alternatives compared to Qwen3-4B-Instruct. Interestingly, after RL training, the smaller model achieves higher scores across most confidence bins, even though this metric was not directly optimized.
These findings highlight the importance of task-specific training for existing LLMs.

\subsection{Discussion on Limitations}
\label{sec:discuss}

\input{tables/math_result}

The premise of predicting a distribution is that the LLM is aware of multiple plausible candidates, and uncertainty can arise from their indistinguishability. However, for certain tasks, predicting multiple answers is inherently difficult. Math reasoning is a typical example: LLMs reason step-by-step to reach a final answer, where the correctness of each step is usually absolute, leaving little room for alternative outcomes. This contrasts sharply with more open-ended tasks like question answering, where multiple answers can be plausible for an LLM (e.g., ``What was Michael Jackson's first album?'').
To validate this, we conducted an experiment on the textual modality portion of the OlympiadBench~\cite{he2024olympiadbench} using Qwen3-4B-Instruct. The training and evaluation details are shown in \S \ref{app:train}.
Results in Table~\ref{tab:math_result} show that both Verbalized Top-$k$ and Verbalized Probability Distribution suffer significant performance drop compared to Verbalized Confidence. However, we demonstrate that RL can largely recover this performance. Another promising approach is to automatically select the proper prompting strategy for different tasks, which we leave for future work.

\section{Related Work}
Early research on model confidence primarily focused on traditional classification tasks, often by adjusting the logits from the classification head~\cite{guptalanguage}. Pioneering techniques such as Histogram Binning~\cite{zadrozny2001obtaining} and Platt Scaling~\cite{platt1999probabilistic} were developed for binary classification. An extension for multi-class settings is Temperature Scaling~\cite{guo2017calibration}, which uses a single scalar parameter to adjust the softmax distribution.

With the advent of LLMs, the research paradigm has shifted toward confidence estimation through prompting. For instance, \citet{kadavath2022language} demonstrated that LLMs can produce well-calibrated confidence scores by prompting them for a true/false determination. In response to the black-box nature of many LLM services and the degradation of calibration due to post-training, verbalization-based approaches have emerged as an effective solution. For example, \citet{tian2023just} found that having the model generate its confidence in natural language yields better calibration. A similar technique involves generating linguistic expressions (e.g., ``highly likely'') and has shown comparable performance~\cite{linteaching,tian2023just}. \citet{xiongcan} further enhanced this method by aggregating scores from repetitive sampling.
The study most related to our work is \cite{wang2024calibrating}, which was the first to investigate verbalized probability distributions. However, it is limited to closed-set tasks and its performance relative to other verbalization-based methods remains unclear.

The aforementioned studies have been predominantly evaluated on simple classification or question-answering tasks. Recent work has extended this inquiry to more complex domains, finding that chain-of-thought reasoning can improve confidence scores~\cite{yoon2025reasoning,devic2025trace,mei2025reasoning}.
Unlike previous work~\cite{azaria2023internal} that used an external classifier to predict confidence scores from intermediate states, this approach is also transparent, clearly showing how the score is estimated.
Notably, \citet{damani2025beyond} trained LLMs using RL to generate reasoning processes specifically for confidence estimation. 
Despite these advances, the characteristics of an effective rationale remain undefined. Our work partially addresses this gap by comparing verbalization-based methods, highlighting the critical role of probability distribution prediction.

Several concurrent studies are also remotely relevant: \citet{zhang2025verbalized} proposed verbalized sampling to generate diverse responses, while \citet{li2025judging} used verbalized probability distributions for preference rating. In contrast, our work focuses specifically on confidence estimation across a variety of tasks, positioning it as a complementary contribution to this field.

\section{Conclusion}
This work investigates confidence estimation through reasoning using verbalized probability distribution prompting. This approach allows LLMs to better express their belief states and assign confidence to possible answers more carefully. Although conceptually simple, this method incentivizes more effective analysis and produces better confidence scores, which is particularly beneficial for smaller LLMs and challenging tasks. These gains are maintained under RL training, indicating that the method is complementary to reasoning trajectory optimization methods. Further analysis validates the effectiveness of each component in our method and confirms its efficiency. However, we also find that prompt effectiveness varies by task, suggesting that future research can focus on automatic prompt selection or optimization for robust and generalizable confidence estimation.

\section*{Limitations}
As discussed in \S \ref{sec:discuss}, the effectiveness of verbalization-based methods may vary across tasks. Although we demonstrate that RL can address this issue, it cannot be directly applied via simple prompting to existing LLMs that lack specific RL training. Flexibly selecting prompting strategies based on task features presents a promising alternative, but it is non-trivial to automatically identify these features and understand their influence on confidence estimation strategies. 
Another limitation is that, although predicting verbalized probability distributions is more efficient than previous methods when combined with inference-time scaling, it consumes more tokens in one pass (a similar issue is also faced by Verbalized Top-$k$). This study does not explore efficient reasoning methods for confidence estimation. Related work on mathematical reasoning has successfully used RL techniques, such as length penalty rewards~\cite{team2025kimi,yeo2025demystifying}, to improve efficiency. While a similar approach could be applied here, its effectiveness remains to be verified.

\bibliography{custom}

\appendix

\section{Datasets}
\label{app:dataset}
The tasks used in this work are categorized based on whether the answer space is closed-set or open-set.
For the closed-set tasks, we evaluate three multiple-choice tasks requiring reasoning: MedQA~\cite{jin2021disease}, MMLU-Pro~\cite{wang2024mmlu}, and MedXpertQA~\cite{zuomedxpertqa}. The test sets comprise 1,273, 12,032, and 2,450 instances, respectively. MedQA and MedXpertQA are both medical benchmarks, with the latter being notably more difficult. MMLU-Pro offers a comprehensive evaluation across diverse domains, such as physical, business, and law.
For the open-set tasks, we employ HotpotQA~\cite{yang2018hotpotqa} and MedCaseReasoning~\cite{wu2025medcasereasoning}. 
HotpotQA focuses on multi-hop question answering, while MedCaseReasoning comprises clinical diagnosis cases.
We utilize the distractor subset of HotpotQA, from which we sample 1,000 test instances. No supporting facts are provided, because we notice this is too simple for recent LLMs. Regarding MedCaseReasoning, we filter the dataset to include only 644 instances that are solvable by the GPT-5, as the original set contains many cases that are prohibitively difficult for most LLMs.

We use accuracy to evaluate performance across all datasets. For closed-set tasks, we compute accuracy using an exact match. For open-set tasks, we employ LLM-as-a-Judge~\cite{zheng2023judging} to compare the prediction against the ground truth. As the ground truth is given, this task can be easily handled even for open-source small LLMs. We consistently adopt Qwen3-30B-A3B-Instruct for evaluation.
For the math reasoning task involved in \S \ref{sec:discuss}, the correctness of an answer is verified with \texttt{Math-Verify}\footnote{\url{https://github.com/huggingface/Math-Verify}}.

\section{Evaluation Metrics}
\label{app:metric}

The evaluation metrics used in this work for confidence estimation are as follow:
\begin{itemize}[leftmargin=*]
    \item \textbf{AUROC} measures a model's ability to discriminate between positive and negative classes, independent of the classification threshold. It is calculated as the area under the plot of the True Positive Rate (TPR) against the False Positive Rate (FPR). For a set of \( n \) predictions, it can be computed as:
    \[
    \text{AUROC} = \frac{\sum_{i=1}^{n_+} \sum_{j=1}^{n_-} \mathbb{I}(s^+_i > s^-_j)}{n_+ \cdot n_-},
    \]
    where \( s^+_i \) and \( s^-_j \) are the confidence scores for the \(i\)-th positive and \(j\)-th negative instance, respectively, and \(n_+\) and \(n_-\) are the total number of positive and negative instances.
    
    \item \textbf{ECE} quantifies calibration, which is the alignment between predicted confidence and empirical accuracy. It is computed by grouping predictions into \( M \) bins (\(B_1, \dots, B_M\)) based on their confidence scores:
    \[
    \text{ECE} = \sum_{m=1}^{M} \frac{|B_m|}{n} \left|\text{acc}(B_m) - \text{conf}(B_m)\right|,
    \]
    where \( \text{acc}(B_m) \) and \( \text{conf}(B_m) \) denote the accuracy and average confidence within bin \( B_m \). A key limitation of ECE is that it does not assess a model's discriminative power; a model can be perfectly calibrated yet fail to separate the classes.
    
    \item \textbf{Brier Score} simultaneously evaluates both calibration and discrimination. It is defined as the mean squared error between the predicted probabilities and the true labels:
    \[
    \text{Brier Score} = \frac{1}{n} \sum_{i=1}^{n} (\hat{p}_i - y_i)^2,
    \]
    where \( \hat{p}_i \) is the predicted probability for instance \( i \) and \( y_i \in \{0, 1\} \) is the corresponding true label. A lower Brier score indicates better overall performance.
\end{itemize}

\section{Evaluated Models}
\label{app:model}
Our evaluation encompasses a diverse set of LLMs of varying scales. This includes the latest models from the Qwen3 series (Qwen3-4B-Instruct-2507 and Qwen3-30B-A3B-Instruct-2507,~\citealt{yang2025qwen3}), the established Mistral-3.2-24B-Instruct~\cite{jiang2023mistral7b}, and the medical-specific Baichuan-M2-32B-GPTQ-INT4~\cite{dou2025baichuan}. We also tested Llama-3.2-3B-Instruct~\cite{touvron2023llama}. However, due to its limited capability, this model often fails to follow output format instructions. Therefore, we only evaluated it within the RL training setting.
Furthermore, we tested more advanced closed-source models, including GPT-4.1~\cite{achiam2023gpt} and DeepSeek-V3-0324~\cite{liu2024deepseek}. We also evaluated recent LLMs capable of performing long chain-of-thought (CoT) reasoning, such as the open-sourced Qwen3-4B-Thinking-2507 and the closed-sourced GPT-5-Mini. Due to the high cost associated with querying APIs for closed-source LLMs, we evaluated these models on a random sample of up to 1,000 instances per dataset.

\section{Training}
\label{app:train}
Our training is based on the prevalent veRL framework~\cite{sheng2024hybridflow}. During the rollout phase, we sample 512 instances, generating 8 rollouts for each with a temperature of 1. For the policy update phase, we use 128 instances and their corresponding rollouts, with a learning rate of 
$1\times10^{-6}$. This configuration results in 4 training steps per rollout step, balancing the efficiency of the rollouts with the preservation of the on-policy nature of the algorithm.
For the math reasoning task, following prior work~\cite{zeng2025simplerl}, we adopt the MATH dataset~\cite{hendrycks2measuring} for RL training. Due to the already strong performance of Qwen3-4B-Instruct on MATH, to ensure training efficacy on challenging problems, we filter the dataset to include only instances where the model fails in at least one of four sampled responses~\cite{team2025kimi}. This yields a final set of 2,359 training instances.

\input{figures/verbalized_prompt}

\section{Conventional Methods}
\label{app:baseline}
We introduce two conventional yet representative methods that use the model's generated token probabilities as confidence scores: Logit and p(True)~\cite{kadavath2022language}.

The \textbf{Logit} method calculates confidence from the token probabilities that constitute the predicted answer. 
Formally, given an input $x$, a model response $\hat{y}$ is sampled from an LLM. This response comprises a reasoning process $r$ and a predicted answer $\hat{a} = (\hat{a}_1, \hat{a}_2, \ldots, \hat{a}_L)$, where $L=|\hat{a}|$. The confidence score is the product of the conditional probabilities for each token:
\[
\text{Conf}_{\text{Logit}} = \prod_{t=1}^{L} P(\hat{a}_t \mid x, r, \hat{a}_{<t}).
\]

The \textbf{p(True)} method prompts the model to evaluate its own generated response, $\hat{y}$. The confidence score is the probability the model assigns to generating the token ``True'' in response to the judgment prompt $p$:
\[
\text{Conf}_{\text{p(True)}} = P(\text{``True''} \mid x, \hat{y}, p).
\]

\input{figures/judge_prompt}

\section{Prompts}
\label{app:prompts}
We summarize the instructions for verbalization-based methods: Verbalized Confidence (Figure~\ref{fig:verb_conf}), Verbalized Top-$k$ (Figure~\ref{fig:verb_topk}), and Verbalized Probability Distribution (Figure~\ref{fig:verb_distrib}).
The prompt for judging the correctness of a prediction against the gold answer is shown in Figure~\ref{fig:judge_prompt}.

\input{tables/top_k}

\section{The Choice of $k$ for Verbalized Top-$k$}
Following previous work~\cite{tian2023just,tao2024trust}, we adopt $k=2$ in the main text. \citet{tian2023just} experimented with $k=2$ and $4$, showing that increasing $k$ does not consistently improve performance. To further validate this finding, we experiment with different values of $k$ on MMLU-Pro, as shown in Table~\ref{tab:top_k}. Our results demonstrate that different values of $k$ yield competitive performance.

\input{tables/multi_brier}

\section{The Quality of the Full Probability Distribution}
In the main text, we follow common practice by evaluating models based on a single final answer derived from the Verbalized Probability Distribution. Our RL reward function is also designed to align with this objective. In this section, we further assess the quality of the entire predicted distribution using the Multi-Brier score (ranging from 0 to 2). A lower score denotes better performance. As shown in Table~\ref{tab:multi_brier}, RL improves the calibration of the full distribution, not just the probability assigned to the final answer.

\input{tables/main_multi_choice}

\input{tables/main_open_end}

\input{tables/app_think}

\section{Detailed Results of Training-Free Experiments}
\label{app:train_free}
Table~\ref{tab:main_multi_choice} and Table~\ref{tab:main_open_end} present the detailed results for the training-free evaluation of various LLMs across specific tasks. These results are consistent with the findings in the main text, showing that the Verbalized Probability Distribution method typically ranks as the best or runner-up across confidence estimation metrics. Results for LLMs capable of long chain-of-thought (CoT) reasoning are shown separately in Table~\ref{tab:train_free_think}.
We observe that, in most cases, directly using long CoT reasoning hurts confidence estimation performance. This underscores the importance of the specialized training explored in our RL experiments. Nevertheless, these results still demonstrate the overall effectiveness of the Verbalized Probability Distribution approach.

\input{tables/rl_multi_choice}

\input{tables/rl_open_end}

\input{figures/train_step_mmlu}

\section{Detailed Results of RL Training Experiments}
\label{app:train_based}
Table~\ref{tab:rl_multi_choice} and~\ref{tab:rl_open_end} present the training results of various LLMs on each task. The findings largely validate the effectiveness of the Verbalized Probability Distribution approach, achieving competitive accuracy while yielding superior scores in confidence estimation metrics.  
One exception is the Qwen3-8B-Base model on out-of-distribution MedCaseReasoning, which shows relatively lower accuracy. This may be attributed to the fact that this model has not undergone post-training, coupled with a discrepancy between learning to predict a probability distribution and learning to select a single best answer. For instance, we observe that multiple candidates within a given distribution often receive the same confidence score, indicating that the model has not learned to adequately distinguish the most likely option, which affects the final evaluation. This suggests that more diverse training domains or adjusted training objectives could be important for better balancing answer prediction and confidence estimation.  
Despite this, our method maintains strong overall performance in most cases, which aligns with the conclusions drawn in the main text.

\input{tables/ood}

\section{Analyses on Out-of-Distribution Tasks}
Research~\cite{chusft} has demonstrated that RL is more effective than supervised fine-tuning at preserving a model's cross-domain capabilities. Echoing previous work~\cite{damani2025beyond}, we again validate RL's strong cross-domain performance on confidence estimation tasks, regardless of the prompts used. This is most apparent when comparing results in Table~\ref{tab:main_open_end} and~\ref{tab:rl_open_end}: although RL does not significantly improve out-of-domain task accuracy, it consistently enhances the quality of confidence estimation.
Using the broader range of topics in MMLU-Pro, we further examine performance across diverse domains in Table~\ref{tab:auroc_medical_ood}. Results show unique advantages of Verbalized Probability Distribution on both in-domain and out-of-domain tasks, which again supports the conclusion.
Interestingly, the selection of the training set is crucial for out-of-domain performance. As shown in Table~\ref{tab:rl_open_end}, training on HotpotQA effectively improves results on MedCaseReasoning, but the reverse is not true. This may be due to the different reasoning complexities of these tasks, which results in varying transfer capabilities across domains.

\input{tables/rlcr_prompt}

\section{Experiment Results with a More Complex Prompt}
Table~\ref{tab:rlcr_prompt} presents the results obtained using a more complex prompt from~\cite{damani2025beyond}. This prompt incorporates fine-grained guidance, such as considering alternative approaches and enumerating plausible uncertainties. Contrary to what might be expected, this complexity did not yield a performance improvement. This finding reinforces that our performance gain is attributable to the requirement for probability distribution prediction.

\input{tables/ablation_multi_choice}

\input{tables/ablation_open_end}

\section{Detailed Results of the Ablation Study}
Tables~\ref{tab:ablation_multi_choice} and \ref{tab:ablation_open_end} present the ablation study results on closed-set and open-set tasks, respectively. In support of the conclusions in the main text, removing the normalization command results in consistently worse performance on closed-set tasks and on the calibration results for open-set tasks, while it reaches competitive AUROC scores on open-set tasks. For example, removing normalization when using GPT-4.1 achieves a better AUROC score on MedCaseReasoning but a poorer one on HotpotQA. This indicates that the normalization operation may have relatively less negative effect on the discrimination capability. For manual normalization, it only reaches better results when applied to more challenging tasks that LLMs struggle with, such as MedXpertQA. This is because LLMs assign more similar confidence scores to the possible answers for these questions. Manual normalization results in a smoother probability distribution, i.e., lower confidence for the final answer. Thus, it aligns more closely with the empirically low accuracy. As a result, its AUROC scores significantly drop, indicating worse confidence estimation.

\input{figures/case_study}

\section{Case Study}
Figure~\ref{fig:case_study} presents a case from the MMLU-Pro test set, demonstrating that the Verbalized Probability Distribution provides more in-depth confidence analyses. This method carefully evaluates the confidence of each option and assigns probabilities through computation and verification. This rigorous process likely leads to the improved final confidence scores.

\end{document}

%% file: tables/unify_train_free_verb.tex
\begin{table*}[]
\centering
\small
\resizebox{0.85\textwidth}{!}{
\begin{tabular}{lcccccccc}
\toprule
& \multicolumn{4}{c}{\textbf{Closed-Set}} & \multicolumn{4}{c}{\textbf{Open-Set}} \\
\cmidrule(lr){2-5} \cmidrule(lr){6-9}
& ACC\,$\uparrow$ & AUROC\,$\uparrow$ & Brier\,$\downarrow$ & ECE\,$\downarrow$ & ACC\,$\uparrow$ & AUROC\,$\uparrow$ & Brier\,$\downarrow$ & ECE\,$\downarrow$ \\
\midrule
Qwen3-4B-Instruct & \textbf{0.553} & -- & -- & -- & 0.301 & -- & -- & -- \\
\textcolor{gray}{$\hookrightarrow$}~~+ Verbalized Conf. & 0.535 & \underline{0.658} & 0.426 & 0.425 & 0.303 & 0.645 & 0.609 & 0.630 \\
\textcolor{gray}{$\hookrightarrow$}~~+ Verbalized Top-$k$ & 0.545 & 0.655 & 0.392 & 0.393 & \underline{0.310} & 0.640 & \underline{0.539} & \underline{0.570} \\
\highlightrow \textcolor{gray}{$\hookrightarrow$}~~+ Verbalized Distrib. & 0.545 & \textbf{0.670} & \textbf{0.247} & \textbf{0.202} & \textbf{0.324} & \underline{0.681} & \textbf{0.333} & \textbf{0.331} \\
\midrule
Qwen3-30B-A3B-Instruct & \underline{0.638} & -- & -- & -- & 0.421 & -- & -- & -- \\
\textcolor{gray}{$\hookrightarrow$}~~+ Verbalized Conf. & \textbf{0.639} & \underline{0.656} & 0.323 & 0.313 & \underline{0.435} & 0.626 & 0.512 & 0.519 \\
\textcolor{gray}{$\hookrightarrow$}~~+ Verbalized Top-$k$ & 0.637 & 0.645 & \underline{0.314} & \underline{0.302} & \textbf{0.444} & 0.634 & 0.468 & 0.477 \\
\highlightrow \textcolor{gray}{$\hookrightarrow$}~~+ Verbalized Distrib. & 0.635 & \textbf{0.680} & \textbf{0.199} & \textbf{0.171} & 0.434 & \textbf{0.681} & \textbf{0.304} & \textbf{0.274} \\
\midrule
GPT-4.1 & \underline{0.714} & -- & -- & -- & 0.665 & -- & -- & -- \\
\textcolor{gray}{$\hookrightarrow$}~~+ Verbalized Conf. & \textbf{0.720} & 0.648 & 0.263 & 0.263 & \textbf{0.672} & 0.679 & 0.288 & 0.277 \\
\textcolor{gray}{$\hookrightarrow$}~~+ Verbalized Top-$k$ & 0.713 & \underline{0.734} & \underline{0.197} & \underline{0.178} & 0.648 & \underline{0.705} & \underline{0.240} & \underline{0.186} \\
\highlightrow \textcolor{gray}{$\hookrightarrow$}~~+ Verbalized Distrib. & \underline{0.714} & \textbf{0.760} & \textbf{0.163} & \textbf{0.136} & \underline{0.669} & \textbf{0.718} & \textbf{0.210} & \textbf{0.100} \\
\midrule
DeepSeek-V3 & 0.652 & -- & -- & -- & \underline{0.576} & -- & -- & -- \\
\textcolor{gray}{$\hookrightarrow$}~~+ Verbalized Conf. & \textbf{0.663} & 0.689 & 0.264 & 0.241 & \textbf{0.583} & \underline{0.712} & 0.301 & 0.285 \\
\textcolor{gray}{$\hookrightarrow$}~~+ Verbalized Top-$k$ & 0.654 & \underline{0.707} & \underline{0.225} & \underline{0.195} & 0.558 & 0.671 & \underline{0.284} & \underline{0.242} \\
\highlightrow \textcolor{gray}{$\hookrightarrow$}~~+ Verbalized Distrib. & \underline{0.656} & \textbf{0.733} & \textbf{0.178} & \textbf{0.143} & 0.573 & \textbf{0.731} & \textbf{0.215} & \textbf{0.094} \\
\bottomrule
\end{tabular}
}
\caption{Comparison of confidence estimation methods across various LLMs, encompassing both the open-source Qwen3 series~\cite{yang2025qwen3} and the closed-source GPT-4.1~\cite{achiam2023gpt} and DeepSeek-V3-0324~\cite{liu2024deepseek}. The best result for each metric is highlighted in \textbf{bold}, and the runner-up is \underline{underlined}. We report detailed results for more models on each specific task in \S \ref{app:train_free}.}
\label{tab:unify_train_free_verb}
\end{table*}

%% file: figures/calibration.tex
\begin{figure*}[t]
    \centering
    \small
    \begin{subfigure}[b]{0.33\textwidth}
        \centering
        \begin{tikzpicture}
        \begin{axis}[
            xlabel={Confidence},
            ylabel={Accuracy},
            ylabel style={yshift=-8pt},
            xmin=0.1, xmax=1,
            ymin=0, ymax=1,
            xtick={0.2,0.4,0.6,0.8,1},
            ytick={0,0.2,0.4,0.6,0.8,1},
            grid=major,
            grid style={dotted,gray!50},
            width=\textwidth,
            height=0.8\textwidth,
            bar width=10pt,
            enlarge x limits=0.05,
        ]
        
        \addplot[ybar, draw=green!70!black, fill=green!40, opacity=0.8] coordinates {
            (0.1, 0.0384)
            (0.2, 0.08)
            (0.3, 0.2844)
            (0.4, 0.4483)
            (0.5, 0.6)
            (0.6, 0.3641)
            (0.7, 0.4419)
            (0.8, 0.4)
            (0.9, 0.4121)
            (1.0, 0.9838)
        };
        
        \addplot[black, dashed, thick] coordinates {
            (0.1,0.1)
            (0.95,0.95)
        };
        \end{axis}
        \end{tikzpicture}
    \end{subfigure}
    \hfill
    \begin{subfigure}[b]{0.33\textwidth}
        \centering
        \begin{tikzpicture}
        \begin{axis}[
            xlabel={Confidence},
            ylabel={Accuracy},
            ylabel style={yshift=-8pt},
            xmin=0.1, xmax=1,
            ymin=0, ymax=1,
            xtick={0.2,0.4,0.6,0.8,1},
            ytick={0,0.2,0.4,0.6,0.8,1},
            grid=major,
            grid style={dotted,gray!50},
            width=\textwidth,
            height=0.8\textwidth,
            bar width=10pt,
            enlarge x limits=0.05,
        ]
        
        \addplot[ybar, draw=blue, fill=blue!40, opacity=0.8] coordinates {
            (0.1, 0.2173)
            (0.2, 0.2272)
            (0.3, 0.2021)
            (0.4, 0.2)
            (0.5, 0.1666)
            (0.6, 0.2910)
            (0.7, 0.3296)
            (0.8, 0.3603)
            (0.9, 0.3980)
            (1.0, 0.7676)
        };

        \addplot[black, dashed, thick] coordinates {
            (0.1,0.1)
            (0.95,0.95)
        };
        \end{axis}
        \end{tikzpicture}
    \end{subfigure}
    \hfill
        \begin{subfigure}[b]{0.33\textwidth}
        \centering
        \begin{tikzpicture}
        \begin{axis}[
            xlabel={Confidence},
            ylabel={Accuracy},
            ylabel style={yshift=-8pt},
            xmin=0.1, xmax=1,
            ymin=0, ymax=1,
            xtick={0.2,0.4,0.6,0.8,1},
            ytick={0,0.2,0.4,0.6,0.8,1},
            grid=major,
            grid style={dotted,gray!50},
            width=\textwidth,
            height=0.8\textwidth,
            bar width=10pt,
            enlarge x limits=0.05,
        ]
        
        \addplot[ybar, draw=red!70!black, fill=red!40, opacity=0.8] coordinates {
            (0.1, 0.0975)
            (0.2, 0.1643)
            (0.3, 0.2536)
            (0.4, 0.3882)
            (0.5, 0.4493)
            (0.6, 0.5272)
            (0.7, 0.5849)
            (0.8, 0.6058)
            (0.9, 0.7129)
            (1.0, 0.8790)
        };
        
        \addplot[black, dashed, thick] coordinates {
            (0.1,0.1)
            (0.95,0.95)
        };
        \end{axis}
        \end{tikzpicture}
    \end{subfigure}
    \caption{Calibration curves of Qwen3-4B-Instruct on MMLU-Pro when using Verbalized Confidence (Left), Verbalized Top-$k$ (Mid), and Verbalized Probability Distribution (Right).}
    \label{fig:calib_curve}
\end{figure*}

%% file: tables/unify_train_based_verb.tex
\begin{table*}[]
\centering
\small
\resizebox{0.85\textwidth}{!}{
\begin{tabular}{lcccccccc}
\toprule
& \multicolumn{4}{c}{\textbf{Closed-Set}} & \multicolumn{4}{c}{\textbf{Open-Set}} \\
\cmidrule(lr){2-5} \cmidrule(lr){6-9}
& ACC\,$\uparrow$ & AUROC\,$\uparrow$ & Brier\,$\downarrow$ & ECE\,$\downarrow$ & ACC\,$\uparrow$ & AUROC\,$\uparrow$ & Brier\,$\downarrow$ & ECE\,$\downarrow$ \\
\midrule
\multicolumn{9}{c}{\textit{RL from the Llama-3.2-3B-Instruct}} \\
\midrule
RLVR & 0.371 & -- & -- & -- & \underline{0.279} & -- & -- & -- \\
\hdashline
RLCR + Conf. & \textbf{0.379} & \underline{0.600} & \underline{0.320} & \underline{0.308} & 0.275 & 0.678 & 0.289 & 0.299 \\
RLCR + Top-$k$ & 0.371 & 0.579 & 0.340 & 0.340 & 0.277 & \underline{0.709} & \underline{0.228} & \underline{0.206} \\
\highlightrow RLCR + Distrib. & \underline{0.375} & \textbf{0.602} & \textbf{0.302} & \textbf{0.290} & \textbf{0.286} & \textbf{0.715} & \textbf{0.170} & \textbf{0.094} \\
\midrule
\multicolumn{9}{c}{\textit{RL from the Qwen3-4B-Instruct}} \\
\midrule
RLVR & 0.564 & -- & -- & -- & \textbf{0.340} & -- & -- & -- \\
\hdashline
RLCR + Conf. & \underline{0.569} & 0.624 & 0.340 & 0.318 & 0.334 & 0.742 & 0.289 & 0.254 \\
RLCR + Top-$k$ & 0.563 & \underline{0.699} & \underline{0.283} & \underline{0.261} & 0.337 & \textbf{0.752} & \underline{0.213} & \underline{0.153} \\
\highlightrow RLCR + Distrib. & \textbf{0.571} & \textbf{0.705} & \textbf{0.228} & \textbf{0.180} & \underline{0.338} & \underline{0.748} & \textbf{0.189} & \textbf{0.093} \\
\midrule
\multicolumn{9}{c}{\textit{Zero-RL from the Qwen3-8B-Base}} \\
\midrule
RLVR & \underline{0.527} & -- & -- & -- & \underline{0.342} & -- & -- & -- \\
\hdashline
RLCR + Conf. & \textbf{0.529} & \underline{0.672} & 0.296 & 0.276 & \textbf{0.343} & \underline{0.718} & 0.271 & 0.232 \\
RLCR + Top-$k$ & 0.515 & 0.667 & \underline{0.240} & \underline{0.200} & 0.326 & 0.717 & \underline{0.202} & \underline{0.137} \\
\highlightrow RLCR + Distrib. & 0.525 & \textbf{0.680} & \textbf{0.219} & \textbf{0.152} & 0.330 & \textbf{0.725} & \textbf{0.181} & \textbf{0.078} \\
\bottomrule
\end{tabular}
}
\caption{Performance of different methods when combined with RL training. Detailed results are reported in \S \ref{app:train_based}.}
\label{tab:unify_train_based_verb}
\end{table*}

%% file: figures/train_step.tex
\begin{figure*}[t]
    \small
    \begin{subfigure}[b]{0.33\textwidth}
        \centering
        \begin{tikzpicture}
        \begin{axis}[
            width=\textwidth,
            height=0.9\textwidth,
            ylabel style={yshift=-8pt},
            xlabel={Training Step},
            ylabel={Accuracy},
            xticklabel={$\pgfmathparse{4*\tick}\pgfmathprintnumber{\pgfmathresult}$},
            ymin=0.735,
            ymax=0.81,
            scaled ticks=false,
        ]
        \addplot[mark=triangle*, color=green!70!black, mark size=1.8] coordinates {
(10, 0.7431)
(20, 0.7431)
(30, 0.7675)
(40, 0.7628)
(50, 0.7683)
(60, 0.7738)
(70, 0.7800)
(80, 0.7800)
(90, 0.7879)
(100, 0.7777)
(110, 0.7879)
(120, 0.7855)
(130, 0.7800)
(140, 0.7934)
(150, 0.7863)
(160, 0.7848)
(170, 0.7722)
(180, 0.7793)
(190, 0.7879)
        };
        
        \addplot[mark=square*, color=blue!70!white, mark size=1.5] coordinates {
(10, 0.7581)
(20, 0.7651)
(30, 0.7659)
(40, 0.7604)
(50, 0.7706)
(60, 0.7722)
(70, 0.7636)
(80, 0.7800)
(90, 0.7840)
(100, 0.7769)
(110, 0.7808)
(120, 0.7871)
(130, 0.7730)
(140, 0.7871)
(150, 0.7840)
(160, 0.7934)
(170, 0.7895)
(180, 0.7706)
(190, 0.7808)
        };
    
        \addplot[mark=*, color=red!80!black, mark size=1.5] coordinates {
            (10,0.7439)
            (20,0.7510)
            (30,0.7604)
            (40,0.7683)
            (50,0.7722)
            (60,0.7620)
            (70,0.7706)
            (80,0.7690)
            (90,0.7675)
            (100,0.7683)
            (110,0.7832)
            (120,0.7808)
            (130,0.7808)
            (140,0.7785)
            (150,0.7769)
            (160,0.7793)
            (170,0.7958)
            (180,0.7871)
            (190,0.7950)
        };
        \end{axis}
        \end{tikzpicture}
    \end{subfigure}
    \begin{subfigure}[b]{0.33\textwidth}
        \centering
        \begin{tikzpicture}
        \begin{axis}[
            width=\textwidth,
            height=0.9\textwidth,
            ylabel style={yshift=-8pt},
            xlabel={Training Step},
            ylabel={AUROC},
            xticklabel={$\pgfmathparse{4*\tick}\pgfmathprintnumber{\pgfmathresult}$},
            scaled ticks=false,
            legend cell align={left},
        ]
        \addplot[mark=triangle*, color=green!70!black, mark size=1.8] coordinates {
(10,0.6620)  
(20,0.6339)  
(30,0.6638)  
(40,0.6393)  
(50,0.6379)  
(60,0.6760)  
(70,0.6573)  
(80,0.6644)  
(90,0.6882)  
(100,0.6401)  
(110,0.6967)  
(120,0.6791)  
(130,0.6191)  
(140,0.6441)  
(150,0.6262)  
(160,0.6366)  
(170,0.6509)  
(180,0.6753)  
(190,0.6723)
        };
        
        \addplot[mark=square*, color=blue!70!white, mark size=1.5] coordinates {
(10,0.6506)  
(20,0.6419)  
(30,0.6233)  
(40,0.6759)  
(50,0.6680)  
(60,0.6719)  
(70,0.6708)  
(80,0.7038)  
(90,0.6915)  
(100,0.6939)  
(110,0.7001)  
(120,0.6888)  
(130,0.7339)  
(140,0.7114)  
(150,0.7389)  
(160,0.7279)  
(170,0.7412)  
(180,0.7461)  
(190,0.7615)
        };
    
        \addplot[mark=*, color=red!80!black, mark size=1.5] coordinates {
(10, 0.6797)
(20, 0.6698)
(30, 0.7126)
(40, 0.7258)
(50, 0.7123)
(60, 0.7152)
(70, 0.7089)
(80, 0.7335)
(90, 0.7636)
(100, 0.7481)
(110, 0.7600)
(120, 0.7659)
(130, 0.7559)
(140, 0.7728)
(150, 0.7561)
(160, 0.7434)
(170, 0.7518)
(180, 0.7813)
(190, 0.7631)
        };
        \end{axis}
        \end{tikzpicture}
    \end{subfigure}
    \begin{subfigure}[b]{0.33\textwidth}
        \centering
        \begin{tikzpicture}
        \begin{axis}[
            width=\textwidth,
            height=0.9\textwidth,
            ylabel style={yshift=-8pt},
            xlabel={Training Step},
            ylabel={Brier Score},
            scaled ticks=false,
            xticklabel={$\pgfmathparse{4*\tick}\pgfmathprintnumber{\pgfmathresult}$},
            legend cell align={left},
            legend pos={north east}
        ]
        \addplot[mark=triangle*, color=green!70!black, mark size=1.8] coordinates {
(10,0.2316)
(20,0.2311)
(30,0.2049)
(40,0.2082)
(50,0.2056)
(60,0.1926)
(70,0.1928)
(80,0.1880)
(90,0.1728)
(100,0.1866)
(110,0.1706)
(120,0.1743)
(130,0.1792)
(140,0.1682)
(150,0.1746)
(160,0.1650)
(170,0.1707)
(180,0.1645)
(190,0.1595)
        };
        \addlegendentry{Conf.}
        
        \addplot[mark=square*, color=blue!70!white, mark size=1.5] coordinates {
(10,0.2036)  
(20,0.1972)  
(30,0.1928)  
(40,0.1810)  
(50,0.1762)  
(60,0.1748)  
(70,0.1780)  
(80,0.1645)  
(90,0.1582)  
(100,0.1635)  
(110,0.1599)  
(120,0.1558)  
(130,0.1598)  
(140,0.1530)  
(150,0.1535)  
(160,0.1470)  
(170,0.1491)  
(180,0.1554)  
(190,0.1514)
        };
        \addlegendentry{Top-$k$}
    
        \addplot[mark=*, color=red!80!black, mark size=1.5] coordinates {
(10,0.1760)  
(20,0.1753)  
(30,0.1630)  
(40,0.1593)  
(50,0.1591)  
(60,0.1635)  
(70,0.1603)  
(80,0.1572)  
(90,0.1511)  
(100,0.1533)  
(110,0.1480)  
(120,0.1452)  
(130,0.1484)  
(140,0.1456)  
(150,0.1493)  
(160,0.1525)  
(170,0.1425)  
(180,0.1411)  
(190,0.1406)
        };
        \addlegendentry{Distrib.}
        \end{axis}
        \end{tikzpicture}
    \end{subfigure}
    \caption{Comparison of different methods on MedQA test set across different training steps when using Qwen3-4B-Instruct. Out-of-distribution results on MMLU-Pro are shown in Figure~\ref{fig:train_step_mmlu}.}
    \label{fig:train_step}
\end{figure*}

%% file: tables/ablation_study.tex
\begin{table*}[]
\centering
\small
\resizebox{0.9\textwidth}{!}{
\begin{tabular}{lcccccccc}
\toprule
& \multicolumn{4}{c}{\textbf{Closed-Set}} & \multicolumn{4}{c}{\textbf{Open-Set}} \\
\cmidrule(lr){2-5} \cmidrule(lr){6-9}
& ACC\,$\uparrow$ & AUROC\,$\uparrow$ & Brier\,$\downarrow$ & ECE\,$\downarrow$ & ACC\,$\uparrow$ & AUROC\,$\uparrow$ & Brier\,$\downarrow$ & ECE\,$\downarrow$ \\
\midrule
Qwen3-4B-Instruct & \textbf{0.545} & \textbf{0.670} & \underline{0.247} & \underline{0.202} & \textbf{0.324} & \underline{0.681} & \underline{0.333} & \underline{0.331} \\
\textcolor{gray}{$\hookrightarrow$}~~$-$ ``None of the above'' & -- & -- & -- & -- & \underline{0.300} & 0.597 & 0.476 & 0.501 \\
\textcolor{gray}{$\hookrightarrow$}~~$-$ Normalization & \underline{0.541} & 0.629 & 0.393 & 0.391 & 0.296 & \textbf{0.693} & 0.421 & 0.435 \\
\textcolor{gray}{$\hookrightarrow$}~~$+$ Manual Normalization & \underline{0.541} & \underline{0.645} & \textbf{0.199} & \textbf{0.145} & 0.296 & 0.636 & \textbf{0.306} & \textbf{0.302} \\
\midrule
GPT-4.1 & \textbf{0.714} & \textbf{0.760} & \underline{0.163} & \underline{0.136} & \textbf{0.669} & \underline{0.718} & \textbf{0.210} & \textbf{0.100} \\
\textcolor{gray}{$\hookrightarrow$}~~$-$ ``None of the above'' & -- & -- & -- & -- & \underline{0.666} & 0.688 & 0.226 & 0.126 \\
\textcolor{gray}{$\hookrightarrow$}~~$-$ Normalization & \underline{0.697} & 0.680 & 0.236 & 0.223 & 0.653 & \textbf{0.733} & 0.233 & 0.175 \\
\textcolor{gray}{$\hookrightarrow$}~~$+$ Manual Normalization & \underline{0.697} & \underline{0.750} & \textbf{0.159} & \textbf{0.109} & 0.655 & 0.674 & \underline{0.223} & \underline{0.125} \\
\bottomrule
\end{tabular}
}
\caption{The ablation study for our method. Specific results for each dataset are shown in Tables~\ref{tab:ablation_multi_choice} and \ref{tab:ablation_open_end}.}
\label{tab:ablation_study}
\end{table*}

%% file: figures/sampling.tex
\begin{figure*}[t]
    \small
    \begin{subfigure}[b]{0.33\textwidth}
        \centering
        \begin{tikzpicture}
        \begin{axis}[
            width=\textwidth,
            height=0.9\textwidth,
            ylabel style={yshift=-8pt},
            xlabel={\# Tokens ($k$)},
            ylabel={AUROC},
            scaled ticks=false,
            legend cell align={left},
            legend pos={south east}
        ]
        \addplot[mark=x, color=gray, mark size=1.8] coordinates {
(1.65, 0.5420)  
(3.30, 0.6426)  
(4.95, 0.6917)  
(6.60, 0.7101)  
(8.25, 0.7214)  
(9.90, 0.7355)  
(11.55, 0.7293)  
(13.20, 0.7338)  
(14.85, 0.7423)  
(16.50, 0.7476)  
(18.15, 0.7515)  
(19.80, 0.7608)  
(21.45, 0.7650)  
(23.10, 0.7683)  
(24.75, 0.7648)  
(26.40, 0.7723)
        };
        \addlegendentry{Base}

        \addplot[mark=triangle*, color=green!70!black, mark size=1.8] coordinates {
(1.66, 0.7628)
(3.32, 0.6269)
(4.98, 0.6575)
(6.64, 0.6885)
(8.30, 0.7126)
(9.96, 0.7228)
(11.62, 0.7290)
(13.28, 0.7325)
(14.94, 0.7315)
(16.60, 0.7404)
(18.26, 0.7461)
(19.92, 0.7471)
(21.58, 0.7496)
(23.24, 0.7545)
(24.90, 0.7561)
(26.56, 0.7634)
        };
        \addlegendentry{Conf.}
        
        \addplot[mark=square*, color=blue!70!white, mark size=1.5] coordinates {
(2.12, 0.7861)
(4.24, 0.7659)
(6.36, 0.7829)
(8.48, 0.7859)
(10.60, 0.8113)
(12.72, 0.8071)
(14.84, 0.8076)
(16.96, 0.8204)
(19.08, 0.8212)
(21.20, 0.8207)
(23.32, 0.8255)
(25.44, 0.8197)
(27.56, 0.8161)
(29.68, 0.8196)
(31.80, 0.8123)
(33.92, 0.8147)
        };
        \addlegendentry{Top-$k$}
    
        \addplot[mark=*, color=red!80!black, mark size=1.5] coordinates {
(2.58, 0.7935)
(5.16, 0.8233)
(7.74, 0.8281)
(10.32, 0.8187)
(12.9, 0.8188)
(15.48, 0.8202)
(18.06, 0.8208)
(20.64, 0.8288)
(23.22, 0.8266)
(25.8, 0.8316)
(28.38, 0.8315)
(30.96, 0.8295)
(33.54, 0.8322)
(36.12, 0.8271)
(38.7, 0.8241)
(41.28, 0.8190)
        };
        \addlegendentry{Distrib.}
        \end{axis}
        \end{tikzpicture}
    \end{subfigure}
    \begin{subfigure}[b]{0.33\textwidth}
        \centering
        \begin{tikzpicture}
        \begin{axis}[
            width=\textwidth,
            height=0.9\textwidth,
            ylabel style={yshift=-8pt},
            xlabel={\# Tokens ($k$)},
            ylabel={Brier Score},
            scaled ticks=false,
        ]
        \addplot[mark=x, color=gray, mark size=1.8] coordinates {
(1.65, 0.2628)  
(3.30, 0.2178)  
(4.95, 0.2059)  
(6.60, 0.1986)  
(8.25, 0.1948)  
(9.90, 0.1903)  
(11.55, 0.1903)  
(13.20, 0.1885)  
(14.85, 0.1857)  
(16.50, 0.1831)  
(18.15, 0.1814)  
(19.80, 0.1800)  
(21.45, 0.1793)  
(23.10, 0.1784)  
(24.75, 0.1781)  
(26.40, 0.1780)
        };

        \addplot[mark=triangle*, color=green!70!black, mark size=1.8] coordinates {
(1.66, 0.2344)
(3.32, 0.2165)
(4.98, 0.2107)
(6.64, 0.2044)
(8.30, 0.1968)
(9.96, 0.1936)
(11.62, 0.1932)
(13.28, 0.1885)
(14.94, 0.1873)
(16.60, 0.1874)
(18.26, 0.1870)
(19.92, 0.1863)
(21.58, 0.1861)
(23.24, 0.1857)
(24.90, 0.1864)
(26.56, 0.1854)
        };
        
        \addplot[mark=square*, color=blue!70!white, mark size=1.5] coordinates {
(2.12, 0.2359)
(4.24, 0.1674)
(6.36, 0.1634)
(8.48, 0.1634)
(10.6, 0.1578)
(12.72, 0.1576)
(14.84, 0.1575)
(16.96, 0.1560)
(19.08, 0.1550)
(21.2, 0.1552)
(23.32, 0.1549)
(25.44, 0.1559)
(27.56, 0.1564)
(29.68, 0.1556)
(31.8, 0.1567)
(33.92, 0.1569)
        };
    
        \addplot[mark=*, color=red!80!black, mark size=1.5] coordinates {
(2.58, 0.1780)
(5.16, 0.1462)
(7.74, 0.1429)
(10.32, 0.1450)
(12.9, 0.1451)
(15.48, 0.1442)
(18.06, 0.1440)
(20.64, 0.1432)
(23.22, 0.1438)
(25.8, 0.1425)
(28.38, 0.1419)
(30.96, 0.1430)
(33.54, 0.1423)
(36.12, 0.1436)
(38.7, 0.1443)
(41.28, 0.1461)
        };
        \end{axis}
        \end{tikzpicture}
    \end{subfigure}
    \begin{subfigure}[b]{0.33\textwidth}
        \centering
        \begin{tikzpicture}
        \begin{axis}[
            width=\textwidth,
            height=0.9\textwidth,
            ylabel style={yshift=-8pt},
            xlabel={\# Tokens ($k$)},
            ylabel={ECE},
            scaled ticks=false
        ]
        \addplot[mark=x, color=gray, mark size=1.8] coordinates {
(1.65, 0.2634)  
(3.30, 0.2117)  
(4.95, 0.2028)  
(6.60, 0.1923)  
(8.25, 0.1887)  
(9.90, 0.1811)  
(11.55, 0.1767)  
(13.20, 0.1723)  
(14.85, 0.1710)  
(16.50, 0.1668)  
(18.15, 0.1637)  
(19.80, 0.1650)  
(21.45, 0.1657)  
(23.10, 0.1655)  
(24.75, 0.1621)  
(26.40, 0.1653)
        };

        \addplot[mark=triangle*, color=green!70!black, mark size=1.8] coordinates {
(1.66, 0.2243)
(3.32, 0.2027)
(4.98, 0.1995)
(6.64, 0.1925)
(8.30, 0.1889)
(9.96, 0.1831)
(11.62, 0.1831)
(13.28, 0.1755)
(14.94, 0.1698)
(16.60, 0.1709)
(18.26, 0.1691)
(19.92, 0.1676)
(21.58, 0.1663)
(23.24, 0.1678)
(24.90, 0.1681)
(26.56, 0.1667)
        };
        
        \addplot[mark=square*, color=blue!70!white, mark size=1.5] coordinates {
(2.12, 0.2313)
(4.24, 0.0416)
(6.36, 0.0535)
(8.48, 0.0628)
(10.60, 0.0734)
(12.72, 0.0718)
(14.84, 0.0716)
(16.96, 0.0872)
(19.08, 0.0830)
(21.20, 0.0859)
(23.32, 0.0939)
(25.44, 0.0943)
(27.56, 0.0940)
(29.68, 0.0894)
(31.80, 0.0874)
(33.92, 0.0927)
        };
    
        \addplot[mark=*, color=red!80!black, mark size=1.5] coordinates {
(2.58, 0.1041)
(5.16, 0.0450)
(7.74, 0.0316)
(10.32, 0.0263)
(12.90, 0.0353)
(15.48, 0.0356)
(18.06, 0.0267)
(20.64, 0.0258)
(23.22, 0.0226)
(25.80, 0.0272)
(28.38, 0.0238)
(30.96, 0.0214)
(33.54, 0.0212)
(36.12, 0.0188)
(38.70, 0.0226)
(41.28, 0.0259)
        };
        \end{axis}
        \end{tikzpicture}
    \end{subfigure}
    \caption{Results of different methods combined with inference-time scaling on MMLU-Pro test set.}
    \label{fig:consistency_mmlu}
\end{figure*}

%% file: figures/answer_space.tex
\begin{figure*}[t]
    \small
    \begin{subfigure}[b]{0.33\textwidth}
        \centering
        \begin{tikzpicture}
        \begin{axis}[
            width=\textwidth,
            height=0.9\textwidth,
            ylabel style={yshift=-8pt},
            xlabel={Confidence},
            ylabel={\# Answer},
            scaled ticks=false,
            legend cell align={left},
        ]

        \addplot[mark=square, color=teal!80!white, mark size=1.5] coordinates {
(0.0533,2.0556)
(0.1625,2.25)
(0.3842,2.9167)
(0.492,2.4)
(0.6,2.4242)
(0.6991,2.375)
(0.7778,2.3333)
(0.8571,2.1411)
(0.9626,1.8473)
        };

        \addplot[mark=o, color=purple!70!blue, mark size=1.5] coordinates {
(0.0453,2.9178)
(0.17,3.2286)
(0.2909,2.9091)
(0.3611,3.0)
(0.5857,2.5)
(0.6885,2.7692)
(0.78,2.3)
(0.8733,2.4)
(0.9674,2.2331)
        };

        \addplot[mark=*, color=blue!80!cyan, mark size=1.5] coordinates {
(0.0772,3.6598)
(0.1634,3.4853)
(0.2676,3.123)
(0.381,2.869)
(0.4605,2.5789)
(0.576,2.74)
(0.6851,2.6081)
(0.7677,2.3548)
(0.8664,2.4754)
(0.9496,2.061)
        };

        \end{axis}
        \end{tikzpicture}
    \end{subfigure}
    \begin{subfigure}[b]{0.33\textwidth}
        \centering
        \begin{tikzpicture}
        \begin{axis}[
            width=\textwidth,
            height=0.9\textwidth,
            ylabel style={yshift=-8pt},
            xlabel={Confidence},
            ylabel={NOA Prob.},
            scaled ticks=false,
            legend cell align={left},
            legend style={font=\scriptsize}
        ]
        \addplot[mark=square, color=teal!80!white, mark size=1.5] coordinates {
(0.0533,0.8944)
(0.1625,0.725)
(0.3842,0.1142)
(0.492,0.117)
(0.6,0.093)
(0.6991,0.0979)
(0.7778,0.065)
(0.8571,0.0624)
(0.9626,0.0228)
        };
        \addlegendentry{GPT-4.1}

        \addplot[mark=o, color=purple!70!blue, mark size=1.5] coordinates {
(0.0474,0.8832)
(0.17,0.6614)
(0.2909,0.4909)
(0.3611,0.0843)
(0.5857,0.0729)
(0.6885,0.12)
(0.78,0.066)
(0.8733,0.045)
(0.9674,0.0177)
        };
        \addlegendentry{Qwen3-4B}
        
        \addplot[mark=*, color=blue!80!cyan, mark size=1.5] coordinates {
(0.0772,0.7264)
(0.1634,0.6003)
(0.2676,0.4381)
(0.381,0.3696)
(0.576,0.207)
(0.6851,0.1369)
(0.7677,0.1129)
(0.8664,0.0639)
(0.9496,0.0124)
        };
        \addlegendentry{Qwen3-4B$+$RL}
    
        \end{axis}
        \end{tikzpicture}
    \end{subfigure}
    \begin{subfigure}[b]{0.33\textwidth}
        \centering
        \begin{tikzpicture}
        \begin{axis}[
            width=\textwidth,
            height=0.9\textwidth,
            ylabel style={yshift=-8pt},
            xlabel={Confidence},
            ylabel={Recall},
            scaled ticks=false,
            legend cell align={left},
        ]
        \addplot[mark=square, color=teal!80!white, mark size=1.5] coordinates {
(0.0533,0.3704)
(0.1625,0.3333)
(0.3842,0.25)
(0.492,0.5)
(0.6,0.697)
(0.6991,0.5446)
(0.7778,0.8611)
(0.8571,0.6805)
(0.9626,0.9142)
        };

        \addplot[mark=o, color=purple!70!blue, mark size=1.5] coordinates {
(0.0453,0.2548)
(0.17,0.5143)
(0.2909,0.3636)
(0.3611,0.2857)
(0.5857,0.2143)
(0.6885,0.6154)
(0.78,0.6)
(0.8733,0.4667)
(0.9674,0.6123)
        };
        
        \addplot[mark=*, color=blue!80!cyan, mark size=1.5] coordinates {
(0.0772,0.1642)
(0.1634,0.3235)
(0.2676,0.541)
(0.381,0.5714)
(0.4605,0.7368)
(0.576,0.82)
(0.6851,0.8108)
(0.7677,0.9677)
(0.8664,0.9344)
(0.9496,0.9634)
        };

        \end{axis}
        \end{tikzpicture}
    \end{subfigure}
    \caption{Predicted candidate answer counts, ``None of the above'' (NOA) probability, and recall of the gold answer as model confidence increases. For robustness, confidence bins containing fewer than 5 instances are excluded.}
    \label{fig:answer_space}
\end{figure*}

%% file: tables/math_result.tex
\begin{table}[t]
\centering
\small
\resizebox{0.48\textwidth}{!}{
\begin{tabular}{lcccc}
\toprule
& ACC\,$\uparrow$ & AUROC\,$\uparrow$ & Brier\,$\downarrow$ & ECE\,$\downarrow$ \\
\midrule
Conf. & 0.734 & 0.741 & 0.240 & 0.239 \\
\textcolor{gray}{$\hookrightarrow$}~~+ RLCR & \underline{0.737} & \underline{0.776} & 0.225 & 0.229 \\
\midrule
Top-$k$ & 0.684 & 0.774 & 0.271 & 0.281 \\
\textcolor{gray}{$\hookrightarrow$}~~+ RLCR & 0.729 & \textbf{0.786} & \textbf{0.196} & \textbf{0.201} \\
\midrule
Distrib. & 0.670 & 0.653 & 0.298 & 0.305 \\
\textcolor{gray}{$\hookrightarrow$}~~+ RLCR & \textbf{0.740} & 0.775 & \underline{0.201} & \underline{0.205} \\
\bottomrule
\end{tabular}
}
\caption{Test results of Qwen3-4B-Instruct on OlympiadBench.}
\label{tab:math_result}
\end{table}

%% file: figures/verbalized_prompt.tex
\begin{figure}[h]
\begin{tcolorbox}[
    title=Prompt for Verbalized Confidence,
    fonttitle=\bfseries\small,
    fontupper=\small,
]
\textbf{Question:} \texttt{[Question]}
\\\\
Reason step-by-step to formulate your final answer. Your answer must be a single entity, a short phrase, or a yes/no. Then, reason about the confidence in your answer. Conclude by providing a JSON object that states the final answer and your estimated confidence in it:
\begin{verbatim}
{
 "final_answer": "Your final answer",
 "confidence": "0-1"
}
\end{verbatim}
\end{tcolorbox}
\caption{The prompt for the Verbalized Confidence method for HotpotQA.}
\label{fig:verb_conf}
\end{figure}

\begin{figure}[h]
\begin{tcolorbox}[
    title=Prompt for Verbalized Top-$k$,
    fonttitle=\bfseries\small,
    fontupper=\small,
]
\textbf{Question:} \texttt{[Question]}
\\\\
Reason step-by-step to formulate 2 best guesses and probability that each is correct. Each answer must be a single entity, a short phrase, or a yes/no.
Your final output must be a JSON array:
\begin{verbatim}
[
{
 "candidate": "first most likely answer",
 "confidence": "0-1"
},
{
 "candidate": "second most likely answer",
 "confidence": "0-1"
}
]
\end{verbatim}
\end{tcolorbox}
\caption{The prompt for the Verbalized Top-$k$ method for HotpotQA.}
\label{fig:verb_topk}
\end{figure}

\begin{figure}[h]
\begin{tcolorbox}[
    title=Prompt for Verbalized Probability Distribution,
    fonttitle=\bfseries\small,
    fontupper=\small,
]
\textbf{Question:} \texttt{[Question]}
\\\\
Reason step-by-step to formulate your answer. You may propose multiple possible answers (fewer than five). Each answer must be a single entity, a short phrase, or a yes/no. Always include "None of the above" as a possible answer. Reason about the confidence in each possible answer. Your final output must be a JSON array where the confidence scores form a probability distribution (they must sum to 1.0):
\begin{verbatim}
[
  {
    "candidate": "Candidate 1",
    "confidence": "0-1"
  },
  {
    "candidate": "Candidate 2",
    "confidence": "0-1"
  },
  ...
  {
    "candidate": "None of the above",
    "confidence": "0-1"
  }
]
\end{verbatim}
\end{tcolorbox}
\caption{The prompt for the Verbalized Probability Distribution method for HotpotQA.}
\label{fig:verb_distrib}
\end{figure}

%% file: figures/judge_prompt.tex
\begin{figure}[h]
\begin{tcolorbox}[
    colback=green!2!white,
    colframe=green!60!black,
    title=Evaluation Prompt for Verification,
    fonttitle=\bfseries\small,
    fontupper=\small,
]
\textbf{Question:} \texttt{[Question]}

\textbf{Ground Truth:} \texttt{[Gold\_Answer]}

\textbf{Prediction:} \texttt{[Predicted\_Answer]}

Is the prediction correct? You must answer ``yes'' or ``no'' only.
\end{tcolorbox}
\caption{The prompt for evaluating the correctness of a predicted answer for HotpotQA.}
\label{fig:judge_prompt}
\end{figure}

%% file: tables/top_k.tex
\begin{table}[t]
\centering
\small
\resizebox{0.47\textwidth}{!}{
\begin{tabular}{lcccc}
\toprule
& ACC\,$\uparrow$ & AUROC\,$\uparrow$ & Brier\,$\downarrow$ & ECE\,$\downarrow$ \\
\midrule
Qwen3-4B-Instruct  & 0.720 & -- & -- & -- \\
\textcolor{gray}{$\hookrightarrow$}~~+ Verbalized Top-2  & 0.716 & 0.772 & 0.236 & 0.230 \\
\textcolor{gray}{$\hookrightarrow$}~~+ Verbalized Top-4  & 0.712 & 0.743 & 0.240 & 0.233 \\
\textcolor{gray}{$\hookrightarrow$}~~+ Verbalized Top-$x$  & 0.695 & 0.749 & 0.231 & 0.216 \\
\bottomrule
\end{tabular}
}
\caption{Results of Verbalized Top-$k$ when using different $k$ on MMLU test set. $x$ is the exact option number of the corresponding question.}
\label{tab:top_k}
\end{table}

%% file: tables/multi_brier.tex
\begin{table}[t]
\centering
\small
\resizebox{0.48\textwidth}{!}{
\begin{tabular}{lccc}
\toprule
& MedQA & MMLU-Pro & MedXpertQA \\
\midrule
Verbalized Distrib.  & 0.387 & 0.452 & 1.190 \\
\textcolor{gray}{$\hookrightarrow$}~~+ RLCR & 0.347 & 0.435 & 1.124 \\
\bottomrule
\end{tabular}
}
\caption{Multi-Label Brier Scores of Qwen3-4B-Instruct on MedQA, MMLU-Pro, MedXpertQA test sets.}
\label{tab:multi_brier}
\end{table}

%% file: tables/main_multi_choice.tex
\begin{table*}[t]
\centering
\small
\resizebox{\textwidth}{!}{
\begin{tabular}{lcccccccccccc}
\toprule
& \multicolumn{4}{c}{\textbf{MedQA}} & \multicolumn{4}{c}{\textbf{MMLU-Pro}} & \multicolumn{4}{c}{\textbf{MedXpertQA}} \\
\cmidrule(lr){2-5} \cmidrule(lr){6-9} \cmidrule(lr){10-13}
& ACC\,$\uparrow$ & AUROC\,$\uparrow$ & Brier\,$\downarrow$ & ECE\,$\downarrow$ & ACC\,$\uparrow$ & AUROC\,$\uparrow$ & Brier\,$\downarrow$ & ECE\,$\downarrow$ & ACC\,$\uparrow$ & AUROC\,$\uparrow$ & Brier\,$\downarrow$ & ECE\,$\downarrow$ \\
\midrule
Qwen3-4B-Instruct & \textbf{0.764} & -- & -- & -- & \textbf{0.720} & -- & -- & -- & \textbf{0.176} & -- & -- & -- \\
\textcolor{gray}{$\hookrightarrow$}~~+ Logit & \textbf{0.764} & \textbf{0.675} & 0.223 & 0.219 & \textbf{0.720} & 0.744 & 0.261 & 0.259 & \textbf{0.176} & 0.505 & 0.810 & 0.808 \\
\textcolor{gray}{$\hookrightarrow$}~~+ p(True) & \textbf{0.764} & 0.642 & 0.215 & 0.199 & \textbf{0.720} & 0.535 & 0.334 & 0.323 & \textbf{0.176} & 0.497 & \underline{0.534} & \underline{0.572} \\
\textcolor{gray}{$\hookrightarrow$}~~+ Verbalized Conf. & 0.735 & 0.663 & 0.238 & 0.224 & 0.711 & 0.764 & 0.250 & 0.243 & 0.160 & \underline{0.548} & 0.789 & 0.808 \\
\textcolor{gray}{$\hookrightarrow$}~~+ Verbalized Top-$k$ & 0.749 & \textbf{0.675} & \underline{0.207} & \underline{0.181} & 0.716 & \underline{0.772} & \underline{0.236} & \underline{0.230} & 0.169 & 0.518 & 0.733 & 0.768 \\
\highlightrow \textcolor{gray}{$\hookrightarrow$}~~+ Verbalized Distrib. & 0.756 & 0.672 & \textbf{0.173} & \textbf{0.044} & 0.707 & \textbf{0.773} & \textbf{0.174} & \textbf{0.091} & 0.172 & \textbf{0.564} & \textbf{0.395} & \textbf{0.472} \\
\midrule
Qwen3-30B-A3B-Instruct & \underline{0.869} & -- & -- & -- & \textbf{0.797} & -- & -- & -- & 0.247 & -- & -- & -- \\
\textcolor{gray}{$\hookrightarrow$}~~+ Logit & \underline{0.869} & \underline{0.683} & 0.123 & 0.121 & \textbf{0.797} & 0.734 & 0.191 & 0.190 & 0.247 & 0.508 & 0.734 & 0.732 \\
\textcolor{gray}{$\hookrightarrow$}~~+ p(True) & \underline{0.869} & \textbf{0.713} & 0.127 & 0.127 & \textbf{0.797} & 0.684 & 0.191 & 0.189 & 0.247 & 0.510 & 0.734 & 0.736 \\
\textcolor{gray}{$\hookrightarrow$}~~+ Verbalized Conf. & 0.866 & 0.679 & \underline{0.117} & \underline{0.088} & 0.794 & 0.724 & 0.179 & 0.162 & \textbf{0.258} & \underline{0.565} & 0.672 & 0.689 \\
\textcolor{gray}{$\hookrightarrow$}~~+ Verbalized Top-$k$ & \textbf{0.870} & 0.655 & \textbf{0.112} & \textbf{0.072} & 0.790 & \underline{0.755} & \underline{0.172} & \underline{0.152} & \underline{0.252} & 0.525 & \underline{0.658} & \underline{0.683} \\
\highlightrow \textcolor{gray}{$\hookrightarrow$}~~+ Verbalized Distrib. & 0.867 & 0.680 & 0.118 & 0.096 & 0.786 & \textbf{0.777} & \textbf{0.140} & \textbf{0.059} & 0.251 & \textbf{0.582} & \textbf{0.339} & \textbf{0.359} \\
\midrule
Mistral-3.2-24B-Instruct & \textbf{0.789} & -- & -- & -- & \textbf{0.666} & -- & -- & -- & \textbf{0.200} & -- & -- & -- \\
\textcolor{gray}{$\hookrightarrow$}~~+ Logit & \textbf{0.789} & 0.697 & 0.194 & 0.196 & \textbf{0.666} & 0.658 & 0.299 & 0.294 & \textbf{0.200} & 0.526 & 0.753 & 0.763 \\
\textcolor{gray}{$\hookrightarrow$}~~+ p(True) & \textbf{0.789} & \textbf{0.777} & \textbf{0.147} & 0.083 & \textbf{0.666} & \textbf{0.735} & \textbf{0.196} & \textbf{0.092} & \textbf{0.200} & 0.539 & 0.520 & \underline{0.576} \\
\textcolor{gray}{$\hookrightarrow$}~~+ Verbalized Conf. & 0.780 & 0.666 & 0.165 & 0.099 & 0.648 & 0.700 & 0.263 & 0.242 & 0.181 & 0.536 & 0.640 & 0.695 \\
\textcolor{gray}{$\hookrightarrow$}~~+ Verbalized Top-$k$ & 0.780 & \underline{0.705} & 0.158 & \textbf{0.047} & 0.657 & 0.721 & 0.229 & 0.180 & 0.186 & \textbf{0.556} & \underline{0.514} & 0.600 \\
\highlightrow \textcolor{gray}{$\hookrightarrow$}~~+ Verbalized Distrib. & 0.785 & 0.699 & \underline{0.155} & \underline{0.052} & 0.659 & \underline{0.723} & \underline{0.209} & \underline{0.129} & \textbf{0.200} & \underline{0.549} & \textbf{0.390} & \textbf{0.444} \\
\midrule
Baichuan-M2-32B-GPTQ-INT4 & \underline{0.823} & -- & -- & -- & \textbf{0.709} & -- & -- & -- & \underline{0.207} & -- & -- & -- \\
\textcolor{gray}{$\hookrightarrow$}~~+ Logit & \underline{0.823} & 0.665 & 0.176 & 0.176 & \textbf{0.709} & 0.560 & 0.287 & 0.288 & \underline{0.207} & 0.481 & 0.788 & 0.789 \\
\textcolor{gray}{$\hookrightarrow$}~~+ p(True) & \underline{0.823} & 0.644 & \underline{0.140} & 0.078 & \textbf{0.709} & 0.520 & 0.248 & 0.151 & \underline{0.207} & 0.459 & 0.570 & 0.617 \\
\textcolor{gray}{$\hookrightarrow$}~~+ Verbalized Conf. & \textbf{0.829} & 0.656 & \textbf{0.130} & \underline{0.043} & 0.701 & \textbf{0.739} & \underline{0.203} & \underline{0.147} & \textbf{0.211} & \underline{0.516} & 0.608 & 0.660 \\
\textcolor{gray}{$\hookrightarrow$}~~+ Verbalized Top-$k$ & 0.791 & \textbf{0.704} & 0.151 & \textbf{0.029} & 0.700 & 0.720 & 0.210 & 0.156 & 0.187 & 0.509 & \underline{0.526} & \underline{0.607} \\
\highlightrow \textcolor{gray}{$\hookrightarrow$}~~+ Verbalized Distrib. & 0.814 & \underline{0.700} & 0.161 & 0.148 & 0.698 & \underline{0.734} & \textbf{0.185} & \textbf{0.070} & \underline{0.207} & \textbf{0.539} & \textbf{0.303} & \textbf{0.345} \\
\midrule
GPT-4.1 & 0.936 & -- & -- & -- & \textbf{0.808} & -- & -- & -- & 0.397 & -- & -- & -- \\
\textcolor{gray}{$\hookrightarrow$}~~+ Verbalized Conf. & 0.936 & 0.679 & 0.059 & \underline{0.053} & \underline{0.806} & 0.715 & 0.172 & 0.172 & \textbf{0.417} & 0.551 & 0.559 & 0.564 \\
\textcolor{gray}{$\hookrightarrow$}~~+ Verbalized Top-$k$ & \underline{0.939} & \textbf{0.822} & \textbf{0.052} & \textbf{0.041} & 0.796 & \underline{0.798} & \underline{0.141} & \underline{0.092} & \underline{0.404} & \underline{0.583} & \underline{0.399} & \underline{0.402} \\
\highlightrow \textcolor{gray}{$\hookrightarrow$}~~+ Verbalized Distrib. & \textbf{0.940} & \underline{0.812} & \underline{0.056} & 0.073 & 0.802 & \textbf{0.812} & \textbf{0.124} & \textbf{0.045} & 0.400 & \textbf{0.656} & \textbf{0.309} & \textbf{0.290} \\
\midrule
DeepSeek-V3 & 0.880 & -- & -- & -- & 0.770 & -- & -- & -- & 0.307 & -- & -- & -- \\
\textcolor{gray}{$\hookrightarrow$}~~+ Verbalized Conf. & \textbf{0.888} & 0.750 & \textbf{0.089} & \textbf{0.029} & \textbf{0.788} & 0.769 & 0.162 & 0.120 & \textbf{0.313} & 0.548 & 0.542 & 0.573 \\
\textcolor{gray}{$\hookrightarrow$}~~+ Verbalized Top-$k$ & 0.882 & \underline{0.796} & \textbf{0.089} & \underline{0.044} & 0.771 & \underline{0.776} & \underline{0.151} & \underline{0.074} & \underline{0.309} & \underline{0.550} & \underline{0.434} & \underline{0.467} \\
\highlightrow \textcolor{gray}{$\hookrightarrow$}~~+ Verbalized Distrib. & \underline{0.883} & \textbf{0.807} & 0.097 & 0.096 & \textbf{0.788} & \textbf{0.817} & \textbf{0.135} & \textbf{0.059} & 0.297 & \textbf{0.574} & \textbf{0.302} & \textbf{0.275} \\
\bottomrule
\end{tabular}
}
\caption{Comparison of training-free approaches on MedQA, MMLU-Pro, and MedXpertQA test sets.}
\label{tab:main_multi_choice}
\end{table*}

%% file: tables/main_open_end.tex
\begin{table*}[t]
\centering
\small
\resizebox{0.78\textwidth}{!}{
\begin{tabular}{lcccccccccccc}
\toprule
& \multicolumn{4}{c}{\textbf{HotpotQA}} & \multicolumn{4}{c}{\textbf{MedCaseReasoning}} \\
\cmidrule(lr){2-5} \cmidrule(lr){6-9}
& ACC\,$\uparrow$ & AUROC\,$\uparrow$ & Brier\,$\downarrow$ & ECE\,$\downarrow$ & ACC\,$\uparrow$ & AUROC\,$\uparrow$ & Brier\,$\downarrow$ & ECE\,$\downarrow$ \\
\midrule
Qwen3-4B-Instruct & 0.303 & -- & -- & -- & 0.298 & -- & -- & -- \\
\textcolor{gray}{$\hookrightarrow$}~~+ Logit & 0.303 & 0.586 & 0.639 & 0.647 & 0.298 & 0.492 & \underline{0.519} & \underline{0.513} \\
\textcolor{gray}{$\hookrightarrow$}~~+ p(True) & 0.303 & \textbf{0.733} & \underline{0.487} & \underline{0.490} & 0.298 & \textbf{0.651} & 0.661 & 0.665 \\
\textcolor{gray}{$\hookrightarrow$}~~+ Verbalized Conf.  & 0.290 & 0.642 & 0.632 & 0.647 & \textbf{0.315} & 0.647 & 0.585 & 0.612 \\
\textcolor{gray}{$\hookrightarrow$}~~+ Verbalized Top-$k$  & \underline{0.319} & 0.656 & 0.550 & 0.572 & 0.301 & 0.623 & 0.527 & 0.568 \\
\highlightrow \textcolor{gray}{$\hookrightarrow$}~~+ Verbalized Distrib.  & \textbf{0.336} & \underline{0.714} & \textbf{0.336} & \textbf{0.322} & \underline{0.311} & \underline{0.648} & \textbf{0.330} & \textbf{0.340} \\
\midrule
Qwen3-30B-A3B-Instruct  & 0.437 & -- & -- & -- & 0.404 & -- & -- & -- \\
\textcolor{gray}{$\hookrightarrow$}~~+ Logit  & 0.437 & 0.596 & 0.485 & 0.486 & 0.404 & 0.567 & 0.381 & 0.338 \\
\textcolor{gray}{$\hookrightarrow$}~~+ p(True)  & 0.437 & \textbf{0.702} & \underline{0.460} & \underline{0.462} & 0.404 & \underline{0.595} & 0.561 & 0.563 \\
\textcolor{gray}{$\hookrightarrow$}~~+ Verbalized Conf.  & 0.426 & 0.664 & 0.518 & 0.526 & \underline{0.443} & 0.587 & 0.506 & 0.512 \\
\textcolor{gray}{$\hookrightarrow$}~~+ Verbalized Top-$k$  & \textbf{0.439} & 0.684 & 0.463 & 0.474 & \textbf{0.449} & 0.583 & \underline{0.472} & \underline{0.480} \\
\highlightrow \textcolor{gray}{$\hookrightarrow$}~~+ Verbalized Distrib.  & \underline{0.438} & \underline{0.689} & \textbf{0.318} & \textbf{0.295} & 0.429 & \textbf{0.673} & \textbf{0.289} & \textbf{0.252} \\
\midrule
GPT-4.1  & \textbf{0.686} & -- & -- & -- & 0.644 & -- & -- & -- \\
\textcolor{gray}{$\hookrightarrow$}~~+ Verbalized Conf.  & 0.679 & 0.728 & 0.291 & 0.290 & \textbf{0.665} & 0.629 & 0.284 & 0.263 \\
\textcolor{gray}{$\hookrightarrow$}~~+ Verbalized Top-$k$  & 0.673 & \underline{0.755} & \underline{0.232} & \underline{0.206} & 0.623 & \underline{0.654} & \underline{0.248} & \underline{0.165} \\
\highlightrow \textcolor{gray}{$\hookrightarrow$}~~+ Verbalized Distrib.  & \underline{0.682} & \textbf{0.772} & \textbf{0.208} & \textbf{0.170} & \underline{0.655} & \textbf{0.663} & \textbf{0.211} & \textbf{0.030} \\
\midrule
DeepSeek-V3  & 0.583 & -- & -- & -- & \underline{0.568} & -- & -- & -- \\
\textcolor{gray}{$\hookrightarrow$}~~+ Verbalized Conf.  & \underline{0.591} & \underline{0.750} & 0.293 & 0.287 & \textbf{0.575} & \underline{0.673} & 0.308 & 0.283 \\
\textcolor{gray}{$\hookrightarrow$}~~+ Verbalized Top-$k$  & 0.572 & 0.698 & \underline{0.276} & \underline{0.241} & 0.543 & 0.644 & \underline{0.292} & \underline{0.242} \\
\highlightrow \textcolor{gray}{$\hookrightarrow$}~~+ Verbalized Distrib.  & \textbf{0.594} & \textbf{0.777} & \textbf{0.203} & \textbf{0.116} & 0.551 & \textbf{0.684} & \textbf{0.227} & \textbf{0.071} \\
\bottomrule
\end{tabular}
}
\caption{Comparison of training-free approaches on HotpotQA and MedCaseReasoning test set.}
\label{tab:main_open_end}
\end{table*}

%% file: tables/app_think.tex
\begin{table*}[t]
\centering
\small
\resizebox{\textwidth}{!}{
\begin{tabular}{lcccccccccccc}
\toprule
& \multicolumn{4}{c}{\textbf{MedQA}} & \multicolumn{4}{c}{\textbf{MMLU-Pro}} & \multicolumn{4}{c}{\textbf{MedXpertQA}} \\
\cmidrule(lr){2-5} \cmidrule(lr){6-9} \cmidrule(lr){10-13}
& ACC\,$\uparrow$ & AUROC\,$\uparrow$ & Brier\,$\downarrow$ & ECE\,$\downarrow$ & ACC\,$\uparrow$ & AUROC\,$\uparrow$ & Brier\,$\downarrow$ & ECE\,$\downarrow$ & ACC\,$\uparrow$ & AUROC\,$\uparrow$ & Brier\,$\downarrow$ & ECE\,$\downarrow$ \\
\midrule
Qwen3-4B-Thinking & 0.762 & -- & -- & -- & \underline{0.722} & -- & -- & -- & 0.168 & -- & -- & -- \\
\textcolor{gray}{$\hookrightarrow$}~~+ Verbalized Conf. & 0.765 & 0.615 & 0.197 & 0.165 & \textbf{0.723} & 0.685 & 0.228 & 0.207 & \underline{0.170} & 0.511 & 0.713 & 0.755 \\
\textcolor{gray}{$\hookrightarrow$}~~+ Verbalized Top-$k$ & \textbf{0.777} & \textbf{0.718} & \underline{0.174} & \underline{0.121} & 0.711 & \textbf{0.780} & \underline{0.209} & \underline{0.188} & \textbf{0.176} & \underline{0.513} & \underline{0.636} & \underline{0.698} \\
\highlightrow \textcolor{gray}{$\hookrightarrow$}~~+ Verbalized Distrib. & \underline{0.770} & \underline{0.698} & \textbf{0.163} & \textbf{0.036} & 0.710 & \underline{0.774} & \textbf{0.181} & \textbf{0.128} & 0.168 & \textbf{0.539} & \textbf{0.497} & \textbf{0.584} \\
\midrule
GPT-5-mini & \underline{0.948} & -- & -- & -- & 0.801 & -- & -- & -- & \textbf{0.445} & -- & -- & -- \\
\textcolor{gray}{$\hookrightarrow$}~~+ Verbalized Conf. & 0.944 & 0.835 & \underline{0.048} & \textbf{0.042} & 0.805 & 0.783 & 0.145 & 0.099 & \textbf{0.445} & \underline{0.603} & 0.419 & 0.425 \\
\textcolor{gray}{$\hookrightarrow$}~~+ Verbalized Top-$k$ & \textbf{0.951} & \underline{0.861} & \textbf{0.046} & 0.086 & \textbf{0.818} & \underline{0.808} & \underline{0.123} & \underline{0.062} & 0.437 & 0.597 & \underline{0.358} & \underline{0.345} \\
\highlightrow \textcolor{gray}{$\hookrightarrow$}~~+ Verbalized Distrib. & 0.941 & \textbf{0.862} & 0.051 & \underline{0.081} & \underline{0.812} & \textbf{0.820} & \textbf{0.120} & \textbf{0.043} & 0.433 & \textbf{0.624} & \textbf{0.280} & \textbf{0.211} \\
\midrule
& \multicolumn{4}{c}{\textbf{HotpotQA}} & \multicolumn{4}{c}{\textbf{MedCaseReasoning}} & \multicolumn{4}{c}{\textbf{Average}} \\
\cmidrule(lr){2-5} \cmidrule(lr){6-9} \cmidrule(lr){10-13}
& ACC\,$\uparrow$ & AUROC\,$\uparrow$ & Brier\,$\downarrow$ & ECE\,$\downarrow$ & ACC\,$\uparrow$ & AUROC\,$\uparrow$ & Brier\,$\downarrow$ & ECE\,$\downarrow$ & ACC\,$\uparrow$ & AUROC\,$\uparrow$ & Brier\,$\downarrow$ & ECE\,$\downarrow$ \\
\midrule
Qwen3-4B-Thinking  & 0.299 & -- & -- & -- & \textbf{0.317} & -- & -- & -- & 0.454 & -- & -- & -- \\
\textcolor{gray}{$\hookrightarrow$}~~+ Verbalized Conf.  & \underline{0.317} & 0.765 & \underline{0.333} & \underline{0.385} & \underline{0.315} & 0.639 & 0.509 & 0.546 & \underline{0.458} & 0.643 & 0.396 & 0.412 \\
\textcolor{gray}{$\hookrightarrow$}~~+ Verbalized Top-$k$  & \textbf{0.330} & \underline{0.776} & 0.376 & 0.432 & 0.309 & \textbf{0.659} & \underline{0.426} & \underline{0.471} & \textbf{0.461} & \underline{0.689} & \underline{0.364} & \underline{0.382} \\
\highlightrow \textcolor{gray}{$\hookrightarrow$}~~+ Verbalized Distrib.  & \underline{0.317} & \textbf{0.781} & \textbf{0.203} & \textbf{0.169} & 0.290 & \underline{0.662} & \textbf{0.356} & \textbf{0.394} & 0.451 & \textbf{0.691} & \textbf{0.280} & \textbf{0.262} \\
\midrule
GPT-5-mini & \underline{0.669} & -- & -- & -- & \textbf{0.725} & -- & -- & -- & \underline{0.718} & -- & -- & -- \\
\textcolor{gray}{$\hookrightarrow$}~~+ Verbalized Conf. & \textbf{0.682} & 0.840 & \underline{0.155} & \underline{0.084} & \underline{0.720} & \textbf{0.737} & \textbf{0.192} & 0.120 & \textbf{0.719} & 0.760 & 0.192 & 0.154 \\
\textcolor{gray}{$\hookrightarrow$}~~+ Verbalized Top-$k$ & 0.659 & \textbf{0.870} & 0.174 & 0.137 & 0.702 & 0.702 & \underline{0.195} & \textbf{0.084} & 0.713 & \underline{0.768} & \underline{0.179} & \underline{0.143} \\
\highlightrow \textcolor{gray}{$\hookrightarrow$}~~+ Verbalized Distrib. & 0.668 & \underline{0.862} & \textbf{0.151} & \textbf{0.082} & 0.691 & \underline{0.723} & 0.196 & \underline{0.103} & 0.709 & \textbf{0.778} & \textbf{0.160} & \textbf{0.104} \\
\bottomrule
\end{tabular}
}
\caption{Results on LLMs capable of long chain-of-thought reasoning.}
\label{tab:train_free_think}
\end{table*}

%% file: tables/rl_multi_choice.tex
\begin{table*}[]
\centering
\small
\resizebox{\textwidth}{!}{
\begin{tabular}{lcccccccccccc}
\toprule
& \multicolumn{4}{c}{\textbf{MedQA}} & \multicolumn{4}{c}{\textbf{MMLU-Pro}} & \multicolumn{4}{c}{\textbf{MedXpertQA}} \\
\cmidrule(lr){2-5} \cmidrule(lr){6-9} \cmidrule(lr){10-13}
& ACC\,$\uparrow$ & AUROC\,$\uparrow$ & Brier\,$\downarrow$ & ECE\,$\downarrow$ & ACC\,$\uparrow$ & AUROC\,$\uparrow$ & Brier\,$\downarrow$ & ECE\,$\downarrow$ & ACC\,$\uparrow$ & AUROC\,$\uparrow$ & Brier\,$\downarrow$ & ECE\,$\downarrow$ \\
\midrule
\multicolumn{13}{c}{\textit{RL from the Llama-3.2-3B-Instruct Model}} \\
\midrule
RLVR & \underline{0.631} & -- & -- & -- & 0.354 & -- & -- & -- & 0.127 & -- & -- & -- \\
\hdashline
RLCR + Conf. & \textbf{0.640} & \textbf{0.664} & \textbf{0.221} & \textbf{0.100} & \underline{0.362} & \underline{0.633} & \underline{0.315} & \underline{0.289} & \textbf{0.134} & \underline{0.502} & \underline{0.423} & \underline{0.536} \\
RLCR + Top-$k$ & 0.630 & 0.626 & 0.238 & 0.131 & 0.353 & 0.622 & 0.335 & 0.323 & 0.129 & 0.488 & 0.448 & 0.566 \\
\highlightrow RLCR + Distrib. & 0.618 & \underline{0.652} & \underline{0.235} & \underline{0.125} & \textbf{0.373} & \textbf{0.639} & \textbf{0.290} & \textbf{0.252} & \underline{0.133} & \textbf{0.514} & \textbf{0.382} & \textbf{0.492} \\
\midrule
\multicolumn{13}{c}{\textit{RL from the Qwen3-4B-Instruct Model}} \\
\midrule
RLVR & 0.782 & -- & -- & -- & \textbf{0.719} & -- & -- & -- & 0.192 & -- & -- & -- \\
\hdashline
RLCR + Conf. & \underline{0.788} & 0.672 & 0.160 & 0.091 & 0.713 & 0.668 & 0.214 & 0.174 & \textbf{0.207} & 0.532 & 0.646 & 0.688 \\
RLCR + Top-$k$ & 0.781 & \underline{0.762} & \underline{0.159} & \underline{0.046} & 0.713 & \textbf{0.801} & \underline{0.191} & \underline{0.155} & 0.194 & \underline{0.534} & \underline{0.499} & \underline{0.581} \\
\highlightrow RLCR + Distrib. & \textbf{0.795} & \textbf{0.763} & \textbf{0.141} & \textbf{0.038} & \underline{0.715} & \underline{0.800} & \textbf{0.159} & \textbf{0.052} & \underline{0.204} & \textbf{0.551} & \textbf{0.384} & \textbf{0.451} \\
\midrule
\multicolumn{13}{c}{\textit{Zero-RL from the Qwen3-Base Model}} \\
\midrule
RLVR & \underline{0.765} & -- & -- & -- & \underline{0.646} & -- & -- & -- & \underline{0.170} & -- & -- & -- \\
\hdashline
RLCR + Conf. & \textbf{0.775} & \underline{0.725} & \textbf{0.155} & \textbf{0.047} & \textbf{0.650} & 0.746 & 0.225 & 0.174 & 0.162 & \underline{0.544} & 0.508 & 0.607 \\
RLCR + Top-$k$ & 0.747 & 0.722 & 0.169 & 0.058 & 0.639 & \underline{0.750} & \underline{0.199} & \underline{0.084} & 0.160 & 0.528 & \underline{0.353} & \underline{0.458} \\
\highlightrow RLCR + Distrib. & \underline{0.765} & \textbf{0.733} & \underline{0.158} & \underline{0.053} & 0.639 & \textbf{0.757} & \textbf{0.187} & \textbf{0.018} & \textbf{0.171} & \textbf{0.551} & \textbf{0.312} & \textbf{0.386} \\
\bottomrule
\end{tabular}
}
\caption{Comparison of RL training-based approaches on MedQA, MMLU-Pro, and MedXpertQA test set.}
\label{tab:rl_multi_choice}
\end{table*}

%% file: tables/rl_open_end.tex
\begin{table*}[]
\centering
\small
\resizebox{0.78\textwidth}{!}{
\begin{tabular}{lcccccccccccc}
\toprule
& \multicolumn{4}{c}{\textbf{HotpotQA}} & \multicolumn{4}{c}{\textbf{MedCaseReasoning}} \\
\cmidrule(lr){2-5} \cmidrule(lr){6-9}
& ACC\,$\uparrow$ & AUROC\,$\uparrow$ & Brier\,$\downarrow$ & ECE\,$\downarrow$ & ACC\,$\uparrow$ & AUROC\,$\uparrow$ & Brier\,$\downarrow$ & ECE\,$\downarrow$ \\
\midrule
\multicolumn{9}{c}{\textit{RL from the Llama-3.2-3B-Instruct Model}} \\
\midrule
RLVR  & \underline{0.380} & -- & -- & -- & 0.177 & -- & -- & -- \\
\hdashline
RLCR + Conf.   & 0.368 & \underline{0.779} & 0.196 & 0.125 & \underline{0.182} & 0.577 & 0.381 & 0.473 \\
RLCR + Top-$k$   & 0.370 & 0.773 & \underline{0.187} & \textbf{0.063} & \textbf{0.183} & \underline{0.644} & \underline{0.268} & \underline{0.348} \\
\highlightrow RLCR + Distrib.  & \textbf{0.397} & \textbf{0.784} & \textbf{0.186} & \underline{0.068} & 0.175 & \textbf{0.645} & \textbf{0.153} & \textbf{0.119} \\
\midrule
\multicolumn{13}{c}{\textit{RL from the Qwen3-4B-Instruct Model}} \\
\midrule
RLVR  & 0.355 & -- & -- & -- & \textbf{0.324} & -- & -- & -- \\
\hdashline
RLCR + Conf.  & 0.345 & \underline{0.838} & \underline{0.156} & 0.054 & \underline{0.322} & \underline{0.646} & 0.421 & 0.454 \\
RLCR + Top-$k$  & \textbf{0.362} & 0.827 & 0.160 & \underline{0.045} & 0.312 & \textbf{0.677} & \underline{0.265} & \underline{0.260} \\
\highlightrow RLCR + Distrib.  & \underline{0.358} & \textbf{0.851} & \textbf{0.147} & \textbf{0.029} & 0.318 & 0.644 & \textbf{0.230} & \textbf{0.156} \\
\midrule
\multicolumn{9}{c}{\textit{Zero-RL from the Qwen3-8B-Base Model}} \\
\midrule
RLVR  & 0.407 & -- & -- & -- & \textbf{0.276} & -- & -- & -- \\
\hdashline
RLCR + Conf.   & \textbf{0.422} & \underline{0.794} & 0.177 & \underline{0.040} & \underline{0.264} & \textbf{0.642} & 0.364 & 0.423 \\
RLCR + Top-$k$   & 0.413 & 0.806 & \underline{0.176} & 0.047 & 0.239 & 0.627 & \underline{0.228} & \underline{0.226} \\
\highlightrow RLCR + Distrib.  & \underline{0.415} & \textbf{0.808} & \textbf{0.172} & \underline{0.039} & 0.245 & \textbf{0.642} & \textbf{0.189} & \textbf{0.116} \\
\midrule
\multicolumn{9}{c}{\textit{RL from Qwen3-4B-Instruct using MedCaseReasoning training set}} \\
\midrule
RLVR  & 0.299 & -- & -- & -- & \underline{0.359} & -- & -- & -- \\
\hdashline
RLCR + Conf.  & 0.282 & 0.616 & 0.558 & 0.582 & \textbf{0.370} & 0.619 & 0.359 & 0.355 \\
RLCR + Top-$k$  & \textbf{0.337} & \underline{0.705} & \underline{0.436} & \underline{0.475} & 0.345 & \underline{0.646} & \underline{0.279} & \underline{0.257} \\
\highlightrow RLCR + Distrib.  & \underline{0.334} & \textbf{0.728} & \textbf{0.279} & \textbf{0.244} & 0.357 & \textbf{0.658} & \textbf{0.227} & \textbf{0.104} \\
\bottomrule
\end{tabular}
}
\caption{Comparison of RL training-based approaches on HotpotQA and MedCaseReasoning test set.}
\label{tab:rl_open_end}
\end{table*}

%% file: figures/train_step_mmlu.tex
\begin{figure*}[t]
    \small
    \begin{subfigure}[b]{0.33\textwidth}
        \centering
        \begin{tikzpicture}
        \begin{axis}[
            width=\textwidth,
            height=0.9\textwidth,
            ylabel style={yshift=-8pt},
            xlabel={Training Step},
            ylabel={Accuracy},
            xticklabel={$\pgfmathparse{4*\tick}\pgfmathprintnumber{\pgfmathresult}$},
            ymin=0.65,
            ymax=0.75,
            scaled ticks=false,
        ]
        \addplot[mark=triangle*, color=green!70!black, mark size=1.8] coordinates {
(10, 0.6934)
(20, 0.6943)
(30, 0.7061)
(40, 0.7007)
(50, 0.6982)
(60, 0.7041)
(70, 0.6992)
(80, 0.7021)
(90, 0.6943)
(100, 0.7026)
(110, 0.6938)
(120, 0.7002)
(130, 0.6982)
(140, 0.6963)
(150, 0.6899)
(160, 0.6855)
(170, 0.6914)
(180, 0.6880)
(190, 0.6919)
        };
        
        \addplot[mark=square*, color=blue!70!white, mark size=1.5] coordinates {
(10, 0.7002)
(20, 0.6895)
(30, 0.6885)
(40, 0.6958)
(50, 0.7061)
(60, 0.6992)
(70, 0.7056)
(80, 0.6992)
(90, 0.6934)
(100, 0.7031)
(110, 0.7007)
(120, 0.7031)
(130, 0.7144)
(140, 0.7139)
(150, 0.7046)
(160, 0.7007)
(170, 0.6997)
(180, 0.6895)
(190, 0.6968)
        };
    
        \addplot[mark=*, color=red!80!black, mark size=1.5] coordinates {
(10, 0.6904)  
(20, 0.6934)  
(30, 0.6807)  
(40, 0.6836)  
(50, 0.6895)  
(60, 0.6860)  
(70, 0.6816)  
(80, 0.6958)  
(90, 0.6865)  
(100, 0.6899)  
(110, 0.6899)  
(120, 0.6963)  
(130, 0.6982)  
(140, 0.6938)  
(150, 0.6929)  
(160, 0.6909)  
(170, 0.7085)  
(180, 0.7002)  
(190, 0.6976)
        };
        \end{axis}
        \end{tikzpicture}
    \end{subfigure}
    \begin{subfigure}[b]{0.33\textwidth}
        \centering
        \begin{tikzpicture}
        \begin{axis}[
            width=\textwidth,
            height=0.9\textwidth,
            ylabel style={yshift=-8pt},
            xlabel={Training Step},
            ylabel={AUROC},
            xticklabel={$\pgfmathparse{4*\tick}\pgfmathprintnumber{\pgfmathresult}$},
            scaled ticks=false,
            legend cell align={left},
            legend pos={south west}
        ]
        \addplot[mark=triangle*, color=green!70!black, mark size=1.8] coordinates {
(10, 0.7613)
(20, 0.7588)
(30, 0.7578)
(40, 0.7310)
(50, 0.7532)
(60, 0.7552)
(70, 0.7306)
(80, 0.7564)
(90, 0.7452)
(100, 0.7430)
(110, 0.7615)
(120, 0.7347)
(130, 0.7218)
(140, 0.7434)
(150, 0.7136)
(160, 0.7169)
(170, 0.7018)
(180, 0.6658)
(190, 0.6768)
        };
        \addlegendentry{Conf.}
        
        \addplot[mark=square*, color=blue!70!white, mark size=1.5] coordinates {
(10, 0.7594)
(20, 0.7602)
(30, 0.7632)
(40, 0.7573)
(50, 0.7372)
(60, 0.7558)
(70, 0.7514)
(80, 0.7630)
(90, 0.7694)
(100, 0.7711)
(110, 0.7660)
(120, 0.7784)
(130, 0.7778)
(140, 0.7711)
(150, 0.7878)
(160, 0.7721)
(170, 0.7827)
(180, 0.7808)
(190, 0.7852)
        };
        \addlegendentry{Top-$k$}
    
        \addplot[mark=*, color=red!80!black, mark size=1.5] coordinates {
(10, 0.7699)
(20, 0.7593)
(30, 0.7756)
(40, 0.7714)
(50, 0.7743)
(60, 0.7601)
(70, 0.7830)
(80, 0.7775)
(90, 0.7851)
(100, 0.7877)
(110, 0.7844)
(120, 0.7960)
(130, 0.7898)
(140, 0.7915)
(150, 0.7809)
(160, 0.7993)
(170, 0.7862)
(180, 0.7897)
(190, 0.7882)
        };
        \addlegendentry{Distrib.}
        \end{axis}
        \end{tikzpicture}
    \end{subfigure}
    \begin{subfigure}[b]{0.33\textwidth}
        \centering
        \begin{tikzpicture}
        \begin{axis}[
            width=\textwidth,
            height=0.9\textwidth,
            ylabel style={yshift=-8pt},
            xlabel={Training Step},
            ylabel={Brier Score},
            scaled ticks=false,
            xticklabel={$\pgfmathparse{4*\tick}\pgfmathprintnumber{\pgfmathresult}$},
            legend cell align={left},
            legend pos={north east}
        ]
        \addplot[mark=triangle*, color=green!70!black, mark size=1.8] coordinates {
(10, 0.2644)
(20, 0.2602)
(30, 0.2497)
(40, 0.2605)
(50, 0.2557)
(60, 0.2483)
(70, 0.2587)
(80, 0.2432)
(90, 0.2469)
(100, 0.2392)
(110, 0.2443)
(120, 0.2367)
(130, 0.2383)
(140, 0.2371)
(150, 0.2390)
(160, 0.2359)
(170, 0.2360)
(180, 0.2334)
(190, 0.2262)
        };
        
        \addplot[mark=square*, color=blue!70!white, mark size=1.5] coordinates {
(10, 0.2463)
(20, 0.2497)
(30, 0.2442)
(40, 0.2302)
(50, 0.2268)
(60, 0.2231)
(70, 0.2176)
(80, 0.2167)
(90, 0.2198)
(100, 0.2100)
(110, 0.2119)
(120, 0.2054)
(130, 0.1943)
(140, 0.1965)
(150, 0.1991)
(160, 0.2040)
(170, 0.2030)
(180, 0.2056)
(190, 0.2006)
        };
    
        \addplot[mark=*, color=red!80!black, mark size=1.5] coordinates {
(10, 0.1778)
(20, 0.1798)
(30, 0.1758)
(40, 0.1768)
(50, 0.1736)
(60, 0.1795)
(70, 0.1704)
(80, 0.1695)
(90, 0.1692)
(100, 0.1669)
(110, 0.1673)
(120, 0.1617)
(130, 0.1638)
(140, 0.1636)
(150, 0.1693)
(160, 0.1664)
(170, 0.1648)
(180, 0.1668)
(190, 0.1679)
        };
        \end{axis}
        \end{tikzpicture}
    \end{subfigure}
    \caption{Comparison of different metrics on randomly sampled 2,048 MMLU-Pro test set across different training steps when using Qwen3-4B-Instruct during RL training.}
    \label{fig:train_step_mmlu}
\end{figure*}

%% file: tables/ood.tex
\begin{table*}[t]
\centering
\resizebox{\textwidth}{!}{
\begin{tabular}{lcccccccccccccc}
\toprule
& \multicolumn{4}{c}{\textbf{In-Domain (Medical)}} & \multicolumn{10}{c}{\textbf{Out-of-Domain}} \\
\cmidrule(lr){2-5} \cmidrule(lr){6-15}
& Hlth & Bio & Chem & Psych & Bus & CS & Econ & Eng & Hist & Law & Math & Oth & Phil & Phys \\
\midrule
Prob. & 0.318 & 0.130 & 0.127 & 0.302 & 0.172 & 0.215 & 0.200 & 0.240 & 0.447 & 0.580 & 0.072 & 0.387 & 0.400 & 0.145 \\
RL + Prob. & 0.260 & 0.136 & \textbf{0.111} & 0.239 & 0.154 & 0.207 & 0.186 & 0.199 & 0.341 & 0.512 & 0.071 & 0.321 & 0.299 & 0.124 \\
\hdashline
Top-$k$ & 0.289 & 0.142 & 0.127 & 0.252 & 0.175 & 0.210 & 0.196 & 0.256 & 0.393 & 0.559 & 0.074 & 0.343 & 0.339 & 0.134 \\
RL + Top-$k$ & 0.230 & \underline{0.121} & \underline{0.114} & 0.199 & 0.143 & \textbf{0.166} & 0.164 & 0.218 & 0.316 & 0.408 & \textbf{0.064} & 0.270 & 0.288 & \underline{0.118} \\
\hdashline
Distrib. & \underline{0.211} & 0.122 & 0.123 & \underline{0.195} & \underline{0.131} & 0.192 & \underline{0.160} & \underline{0.190} & \underline{0.271} & \underline{0.322} & \underline{0.066} & \underline{0.240} & \underline{0.240} & \underline{0.118} \\
RL + Distrib. & \textbf{0.190} & \textbf{0.116} & 0.117 & \textbf{0.168} & \textbf{0.128} & \underline{0.171} & \textbf{0.141} & \textbf{0.172} & \textbf{0.231} & \textbf{0.297} & 0.067 & \textbf{0.211} & \textbf{0.224} & \textbf{0.110} \\
\bottomrule
\end{tabular}
}
\caption{Brier Score performance on In-Domain (Medical) vs Out-of-Domain tasks from the MMLU-Pro test set.}
\label{tab:auroc_medical_ood}
\end{table*}

%% file: tables/rlcr_prompt.tex
\begin{table*}[]
\centering
\small
\resizebox{0.8\textwidth}{!}{
\begin{tabular}{lcccccccccccc}
\toprule
& \multicolumn{4}{c}{\textbf{HotpotQA}} & \multicolumn{4}{c}{\textbf{MedCaseReasoning}} \\
\cmidrule(lr){2-5} \cmidrule(lr){6-9}
& ACC\,$\uparrow$ & AUROC\,$\uparrow$ & Brier\,$\downarrow$ & ECE\,$\downarrow$ & ACC\,$\uparrow$ & AUROC\,$\uparrow$ & Brier\,$\downarrow$ & ECE\,$\downarrow$ \\
\midrule
Qwen3-4B-Instruct  & 0.297 & 0.739 & 0.433 & 0.455 & 0.283 & 0.656 & 0.544 & 0.588 \\
\textcolor{gray}{$\hookrightarrow$}~~+ RLCR & 0.332 & 0.799 & 0.168 & 0.051 & 0.310 & 0.665 & 0.319 & 0.339 \\
\hdashline
Qwen3-30B-A3B-Instruct  & 0.433 & 0.659 & 0.455 & 0.464 & 0.396 & 0.612 & 0.489 & 0.498 \\
GPT-4.1  & 0.665 & 0.745 & 0.297 & 0.301 & 0.596 & 0.647 & 0.354 & 0.349 \\
DeepSeek-V3-0324  & 0.558 & 0.744 & 0.285 & 0.268 & 0.523 & 0.624 & 0.358 & 0.343 \\
\bottomrule
\end{tabular}
}
\caption{Results on HotpotQA and MedCaseReasoning test set using the long prompt from~\cite{damani2025beyond}.}
\label{tab:rlcr_prompt}
\end{table*}

%% file: tables/ablation_multi_choice.tex
\begin{table*}[t]
\centering
\small
\resizebox{\textwidth}{!}{
\begin{tabular}{lcccccccccccc}
\toprule
& \multicolumn{4}{c}{\textbf{MedQA}} & \multicolumn{4}{c}{\textbf{MMLU-Pro}} & \multicolumn{4}{c}{\textbf{MedXpertQA}} \\
\cmidrule(lr){2-5} \cmidrule(lr){6-9} \cmidrule(lr){10-13}
& ACC\,$\uparrow$ & AUROC\,$\uparrow$ & Brier\,$\downarrow$ & ECE\,$\downarrow$ & ACC\,$\uparrow$ & AUROC\,$\uparrow$ & Brier\,$\downarrow$ & ECE\,$\downarrow$ & ACC\,$\uparrow$ & AUROC\,$\uparrow$ & Brier\,$\downarrow$ & ECE\,$\downarrow$ \\
\midrule
Qwen3-4B-Instruct & \textbf{0.756} & \textbf{0.672} & \textbf{0.173} & \textbf{0.044} & \textbf{0.707} & \textbf{0.773} & \textbf{0.174} & \textbf{0.091} & 0.172 & \textbf{0.564} & 0.395 & 0.472 \\
\textcolor{gray}{$\hookrightarrow$}~~$-$ Norm. & \textbf{0.756} & 0.642 & 0.208 & 0.180 & 0.682 & 0.715 & 0.250 & 0.239 & \textbf{0.185} & 0.531 & 0.721 & 0.753 \\
\textcolor{gray}{$\hookrightarrow$}~~$+$ Manual Norm. & \textbf{0.756} & 0.632 & 0.188 & 0.097 & 0.682 & 0.750 & 0.192 & 0.124 & \textbf{0.185} & 0.553 & \textbf{0.216} & \textbf{0.215} \\
\midrule
GPT-4.1 & \textbf{0.940} & 0.812 & \textbf{0.056} & 0.073 & \textbf{0.802} & \textbf{0.812} & \textbf{0.124} & \textbf{0.045} & \textbf{0.400} & \textbf{0.656} & 0.309 & 0.290 \\
\textcolor{gray}{$\hookrightarrow$}~~$-$ Norm. & 0.928 & 0.790 & 0.057 & \textbf{0.032} & 0.774 & 0.691 & 0.179 & 0.164 & 0.388 & 0.558 & 0.471 & 0.472 \\
\textcolor{gray}{$\hookrightarrow$}~~$+$ Manual Norm. & 0.928 & \textbf{0.832} & 0.065 & 0.084 & 0.774 & 0.778 & 0.167 & 0.129 & 0.388 & 0.641 & \textbf{0.244} & \textbf{0.115} \\
\bottomrule
\end{tabular}
}
\caption{Ablation study results for our method on MedQA, MMLU-Pro, and MedXpertQA test sets.}
\label{tab:ablation_multi_choice}
\end{table*}

%% file: tables/ablation_open_end.tex
\begin{table*}[t]
\centering
\small
\resizebox{0.78\textwidth}{!}{
\begin{tabular}{lcccccccccccc}
\toprule
& \multicolumn{4}{c}{\textbf{HotpotQA}} & \multicolumn{4}{c}{\textbf{MedCaseReasoning}} \\
\cmidrule(lr){2-5} \cmidrule(lr){6-9}
& ACC\,$\uparrow$ & AUROC\,$\uparrow$ & Brier\,$\downarrow$ & ECE\,$\downarrow$ & ACC\,$\uparrow$ & AUROC\,$\uparrow$ & Brier\,$\downarrow$ & ECE\,$\downarrow$ \\
\midrule
Qwen3-4B-Instruct  & \textbf{0.336} & 0.714 & \textbf{0.336} & \textbf{0.322} & \textbf{0.311} & \textbf{0.648} & 0.330 & 0.340 \\
\textcolor{gray}{$\hookrightarrow$}~~$-$ ``None of the above'' & 0.311 & 0.558 & 0.585 & 0.594 & 0.289 & 0.636 & 0.366 & 0.408 \\
\textcolor{gray}{$\hookrightarrow$}~~$-$ Norm.  & 0.318 & \textbf{0.741} & 0.357 & 0.335 & 0.273 & 0.645 & 0.485 & 0.534 \\
\textcolor{gray}{$\hookrightarrow$}~~$+$ Manual Norm. & 0.319 & 0.679 & 0.350 & 0.358 & 0.273 & 0.593 & \textbf{0.261} & \textbf{0.245} \\
\midrule
GPT-4.1  & \textbf{0.682} & \textbf{0.772} & \textbf{0.208} & 0.170 & 0.655 & 0.663 & 0.211 & \textbf{0.030} \\
\textcolor{gray}{$\hookrightarrow$}~~$-$ ``None of the above'' & 0.671 & 0.715 & 0.241 & 0.211 & \textbf{0.660} & 0.661 & \textbf{0.210} & 0.040\\
\textcolor{gray}{$\hookrightarrow$}~~$-$ Norm. & 0.668 & 0.758 & 0.250 & 0.234 & 0.637 & \textbf{0.708} & 0.215 & 0.115 \\
\textcolor{gray}{$\hookrightarrow$}~~$+$ Manual Norm. & 0.672 & 0.698 & 0.221 & \textbf{0.164} & 0.638 & 0.649 & 0.224 & 0.086 \\
\bottomrule
\end{tabular}
}
\caption{Ablation study results for our method on HotpotQA and MedCaseReasoning test sets.}
\label{tab:ablation_open_end}
\end{table*}

%% file: figures/case_study.tex
\begin{figure*}[t]
\begin{tcolorbox}[
    colback=blue!5!white,
    colframe=blue!75!black,
    title=Case study on MMLU-Pro,
    fonttitle=\bfseries,
    fontupper=\small
]
\textbf{Question:} 
Typical advertising regulatory bodies suggest, for example that adverts must not: encourage \_\_\_\_\_, cause unnecessary \_\_\_\_\_ or \_\_\_\_\_, and must not cause \_\_\_\_\_ offence.

\textbf{Options:}
A. Safe practices, Fear, Jealousy, Trivial
B. Unsafe practices, Distress, Joy, Trivial
C. Safe practices, Wants, Jealousy, Trivial
D. Safe practices, Distress, Fear, Trivial
E. Unsafe practices, Wants, Jealousy, Serious
F. Safe practices, Distress, Jealousy, Serious
G. Safe practices, Wants, Fear, Serious
H. Unsafe practices, Wants, Fear, Trivial
I. Unsafe practices, Distress, Fear, Serious

---------------------------------------------------------------------------------------------------------------------------------------------

\textbf{Verbalized Confidence:}

Let's solve this step by step.
...
The reasoning is based on standard advertising regulation principles. The combination of "unsafe practices", "distress", "fear", and "serious offence" is well-documented in regulatory guidelines. No other option fits all parts correctly.
Therefore, I am highly confident.

Final Answer:
\begin{verbatim}
{"final_answer": "I", "confidence": 0.98}
\end{verbatim}

\textbf{Verbalized Top-$k$:}

Let's analyze the question step by step:
...
Final decision:
Best: I  
Second: H (with lower confidence due to "wants")
But "wants" is not typically regulated.
After re-evaluation, the only option that makes sense is I.
Thus, second best might be B — but "joy" is not a negative.
No.
Therefore, I conclude:
Best: I  
Second: None strong — but if forced, perhaps H.
But let's go with:
- I: 0.95  
- H: 0.35 (due to "wants" being out of place)
Alternatively, perhaps the original intended answer is I.

Final Answer:
\begin{verbatim}
[
  {"option": "I", "confidence": 0.95},
  {"option": "H", "confidence": 0.35}
]
\end{verbatim}

\textbf{Verbalized Probability Distribution:}

Let's analyze the question step by step:
...
1. I $\rightarrow$ Strongly correct $\rightarrow$ High confidence
2. E $\rightarrow$ Close, but ``wants'' is not a standard concern $\rightarrow$ Medium confidence
3. H $\rightarrow$ ``wants'' is problematic $\rightarrow$ Medium-low confidence
4. B $\rightarrow$ ``joy'' is wrong $\rightarrow$ Low confidence
5. D, F, G, C, A $\rightarrow$ contain ``safe practices'' $\rightarrow$ Wrong direction $\rightarrow$ Very low confidence
6. All others $\rightarrow$ flawed in key points $\rightarrow$ Very low

Step 5: Assign Confidence Scores (must sum to 1.0)
I: 0.60
E: 0.15
H: 0.10
B: 0.05
Others (A, C, D, F, G): 0.00 (too wrong)
Wait — we must assign to all options.
Total: 1.0
Better distribution:
I: 0.60
E: 0.15
H: 0.10
B: 0.05
A, C, D, F, G: 0.05 each $\rightarrow$ sum = $0.05 \times 5 = 0.25 \rightarrow$ total = $0.60 + 0.15 + 0.10 + 0.05 + 0.25 = 1.15 \rightarrow$ too high.
Adjust:
I: 0.60
E: 0.15
H: 0.10
B: 0.05
A, C, D, F, G: 0.05 each $\rightarrow 0.05 \times 5 = 0.25 \rightarrow$ total = $0.60 + 0.15 + 0.10 + 0.05 + 0.25 = 1.15 \rightarrow$ still over.
So reduce the others:
Better:
I: 0.60
E: 0.15
H: 0.10
B: 0.05
A: 0.02
C: 0.02
D: 0.02
F: 0.02
G: 0.02
Sum: $0.60 + 0.15 = 0.75$; $+0.10 = 0.85$; $+0.05 = 0.90$; $+0.02 \times 5 = 0.10 \rightarrow$ total = 1.00.
Yes.

Final Confidence Scores:
A: 0.02
B: 0.05
C: 0.02
D: 0.02
E: 0.15
F: 0.02
G: 0.02
H: 0.10
I: 0.60
Wait — H is better than B. Let's reassess.
Actually, H has ``unsafe practices, wants, fear, trivial'' — ``wants'' is weak, ``trivial'' is weaker than ``serious''.
E has ``serious offence'' — better than trivial.
So I is best.
H: ``fear'' is good, ``wants'' is bad $\rightarrow$ lower than E.
So better to give:
I: 0.60
E: 0.15
H: 0.10
B: 0.05
A: 0.02
C: 0.02
D: 0.02
F: 0.02
G: 0.02
Sum: $0.60 + 0.15 + 0.10 + 0.05 + 0.02 \times 5 = 0.60 + 0.15 + 0.10 + 0.05 + 0.10 = 1.00$
Yes.

Final Answer:
\begin{verbatim}
[
  {"option": "A", "confidence": "0.02"},
  {"option": "B", "confidence": "0.05"},
  {"option": "C", "confidence": "0.02"},
  {"option": "D", "confidence": "0.02"},
  {"option": "E", "confidence": "0.15"},
  {"option": "F", "confidence": "0.02"},
  {"option": "G", "confidence": "0.02"},
  {"option": "H", "confidence": "0.10"},
  {"option": "I", "confidence": "0.60"}
]
\end{verbatim}
\end{tcolorbox}
\caption{A test case from the MMLU-Pro test set, comparing different verbalization-based methods.}
\label{fig:case_study}
\end{figure*}